\documentclass[letterpaper, 10 pt, conference]{IEEEtran}
\usepackage{times}

\makeatletter
\let\NAT@parse\undefined
\makeatother

\usepackage{amssymb,amsmath}

\usepackage[sort&compress,numbers,sectionbib]{natbib}
\usepackage{cite} 
\usepackage{booktabs}
\usepackage{subcaption}
\usepackage{graphicx}

\usepackage{verbatimbox}
\usepackage{multirow}
\usepackage{balance}

\newcommand{\figref}[1]{Fig.~\ref{#1}}
\newcommand{\tabref}[1]{Table~\ref{#1}}

\usepackage[hyphens]{url}

\usepackage{hyperref}
\hypersetup{bookmarksopen,bookmarksnumbered,
pdfpagemode=UseOutlines,
colorlinks=true,
linkcolor=blue,
anchorcolor=blue,
citecolor=blue,
filecolor=blue,
menucolor=blue,
urlcolor=blue,
breaklinks=true
}
\usepackage[dvipsnames]{xcolor}
\definecolor{myred}{RGB}{215,48,39}
\definecolor{myblue}{RGB}{69,117,180}
\definecolor{myorange}{RGB}{252,141,89}
\definecolor{mylightblue}{RGB}{145,191,219}

\definecolor{MYlightblue}{RGB}{217,95,2} 
\definecolor{MYdarkblue}{RGB}{117,112,179} 
\definecolor{MYgreen}{RGB}{27,158,119}

\usepackage{color,soul} 
\definecolor{krishna}{rgb}{.91,.65,.60} 
\definecolor{krishna-text}{rgb}{.59,.24,.19} 
\usepackage{svg}

\newcommand{\xxnote}[3]{}
\ifx\hidenotes\undefined
  \usepackage{color}
  \renewcommand{\xxnote}[3]{\color{#2}{#1: #3}}
\fi

\usepackage{todonotes}

\usepackage{enumitem}
\usepackage{bm}


\IEEEoverridecommandlockouts

\usepackage{savesym}
\savesymbol{comment}
\usepackage{changes}
\restoresymbol{CHNG}{comment}
\setaddedmarkup{\color{RoyalBlue}{#1}}
\setdeletedmarkup{\color{red}{\sout{#1}}}

\begin{document}

\title{Winding Through:\\ Crowd Navigation via Topological Invariance}

\author{Christoforos Mavrogiannis$^{1}$, Krishna Balasubramanian$^{2}$, Sriyash Poddar$^{3}$\\ Anush Gandra$^{2}$, Siddhartha S. Srinivasa$^{1}$
\thanks{$^{1}$Paul G. Allen School of Computer Science \& Engineering, University of Washington, Seattle, USA. Email: \texttt{\{cmavro, siddh\}@cs.washington.edu}}
\thanks{$^{2}$Department of Mechanical Engineering, University of Washington, Seattle, USA. Email: \texttt{\{krishnab, agandr\}@uw.edu}.}
\thanks{$^{3}$Department of Computer Science and Engineering, Indian Institute of Technology, Kharagpur, Kharagpur, India. Email: \texttt{poddarsriyash@iitkgp.ac.in}}
\thanks{This work was (partially) funded by the Honda Research Institute USA, the National Science Foundation NRI (\#2132848) and CHS (\#2007011), DARPA RACER (\#HR0011-21-C-0171), the Office of Naval Research (\#N00014-17-1-2617-P00004 and \#2022-016-01 UW), and Amazon.}
}


\maketitle

\begin{abstract}
We focus on robot navigation in crowded environments. The challenge of predicting the motion of a crowd around a robot makes it hard to ensure human safety and comfort. Recent approaches often employ end-to-end techniques for robot control or deep architectures for high-fidelity human motion prediction. While these methods achieve important performance benchmarks in simulated domains, dataset limitations and high sample complexity tend to prevent them from transferring to real-world environments. Our key insight is that a low-dimensional representation that captures critical features of crowd-robot dynamics could be sufficient to enable a robot to wind through a crowd smoothly. To this end, we mathematically formalize the act of \emph{passing} between two agents as a rotation, using a notion of topological invariance. Based on this formalism, we design a cost functional that favors robot trajectories contributing higher passing progress and penalizes switching between different sides of a human. We incorporate this functional into a model predictive controller that employs a simple constant-velocity model of human motion prediction. This results in robot motion that accomplishes statistically significantly higher clearances from the crowd compared to state-of-the-art baselines while maintaining competitive levels of efficiency, across extensive simulations and challenging real-world experiments on a self-balancing robot.

\end{abstract}




\IEEEpeerreviewmaketitle

\section{Introduction}\label{sec:intro}

Navigation in crowds is a challenging task for a mobile robot~\citep{core-challenges2021}. Ensuring human safety~\citep{salvini-billard} and comfort requires equipping robots with models of human motion prediction. Predicting the motion of a crowd is fundamentally hard due to the combinatorial structure of the space~\citep{cooper90}. This hardness is also reflected in the high sample complexity of data-driven models for human motion prediction~\citep{rudenko2019-predSurvey}. Crucially, predicting human motion alongside a robot can be even harder: human response to robot motion is not well understood, and often driven by novelty effects, limited human mental models, and context dependency~\citep{rae13,joosse21,reinhardt21,walker2021corl}. Collecting demonstrations at scale in such domains is complicated and existing datasets are still limited.

Despite these challenges, recent approaches to robot navigation in crowds overemphasize end-to-end control policies or data-driven models for human trajectory prediction~\citep{chen-icra17,Everett18_IROS,chen2019crowd,liu2020decentralized}. While these approaches successfully handle controlled or simulated domains, they suffer from data limitations, exacerbated by their high sample complexity and the lack of reliable crowd simulation engines, exhibiting poor generalization and transfer to real-world domains. 

Interestingly, as shown by~\citet{aravantinos}, constant velocity (CV) prediction, a very coarse human motion model may outperform even complex neural architectures on prediction benchmarks across common pedestrian datasets. This can be due to the tendency of neural networks to model model environmental priors, the irrelevance of pedestrians' motion history or the hardness of modeling multiagent interactions~\citep{cooper90}. 


\begin{figure}[t!]
    \centering
    \includegraphics[width = \columnwidth]{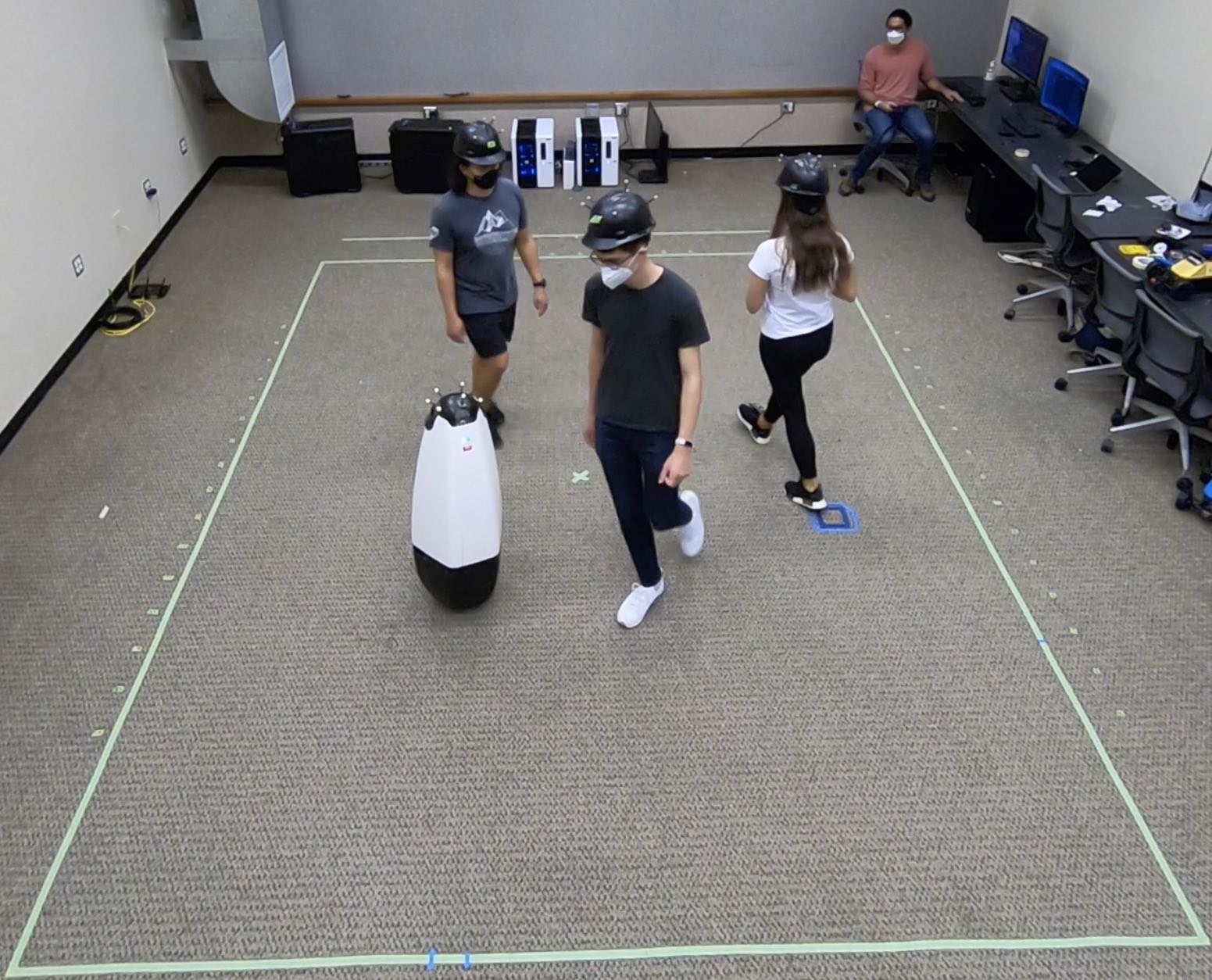}
    \caption{Honda's experimental ballbot~\citep{pathbot} navigates next to three users during our lab experiments.}
    \label{fig:introfig}
\end{figure}



While the CV model indeed handles some prediction tasks, its inability to model cooperation, the foundation of social order in human navigation~\citep{Wolfinger95}, results in notable crowd navigation issues like the ``freezing robot'' problem~\citep{trautmanijrr} and the ``reciprocal dance"~\citep{feurtey2000simulating}. However, encouraged by its efficacy, our key insight is that a coarse motion model like CV could be sufficient for a controller that models critical aspects of interaction. Following this insight, we describe a low-dimensional abstraction of multiagent dynamics based on a formalism of pairwise \emph{passing}, capturing the aspects of progress and directionality using a notion of topological invariance. Based on this abstraction, we design a cost that favors robot trajectories expediting agile passing maneuvers through a crowd. Around this cost, we design a model predictive control (MPC) architecture that employs CV as the human transition model. Through challenging real-world and simulated scenarios involving navigation of a self-balancing robot in a crowded environment, we demonstrate that our controller exhibits significantly safer behavior for similar efficiency compared to state-of-the-art end-to-end and model-based baselines. A video with snippets from our experiments can be found at \href{https://youtu.be/o1o_Rd7MEcg}{this} link.

\section{Related Work}\label{sec:relatedwork}





We review work on robot navigation in crowds, highlighting approaches that model multiagent cooperation to inform robot decision-making.


\subsection{Robot Navigation in Crowds}

Recent work on crowd navigation has employed data-driven techniques to extract social objectives and prediction models from datasets~\citep{PellegriniESG09,UCY} and crowd simulation engines~\citep{tsoi20,socnavbench}. Several approaches used inverse reinforcement learning (IRL) techniques to recover socially aware control policies. \citet{ziebart2009} learned a reward function that describes human decision making for navigation and used it during planning to avoid obstructing human paths. \citet{kretzschmar_ijrr16} and \citet{Kim2016} learned robot reward functions incorporating social preferences like passing from the right side or respecting users' personal space. More recently, \citet{Everett18_IROS,chen2019crowd,liu2020decentralized} employed deep reinforcement learning (RL) to learn control policies for decentralized multiagent collision avoidance. Motivated by real-world constraints, \citet{darpino} employed RL to learn a navigation policy for obstacle-cluttered indoor environments whereas \citet{Monaci-RSS-22} developed an RL vision-based architecture for robots with narrow field-of-view cameras. In parallel, some approaches incorporated data-driven predictions into the general MPC pipeline. For instance, \citet{nishimura2020} and \citet{roh2020corl} integrated multimodal trajectory prediction~\citep{trajectron} into the MPC pipeline whereas \citet{wang2021corl} incorporated group-based behavior prediction to avoid intruding the social space of pedestrian groups. 

Despite the success of these methods in controlled and simulated environments, their sample complexity coupled with the inability of existing crowd simulators to realistically model crowd motion next to robots limits their applicability and generalization to new domains. To mitigate this issue, recent work blends data-driven and model-based reasoning. For instance, \citet{brito-ral21} employed an interaction-aware RL-based subgoal recommendation model to smoothly guide a MPC towards the robot's goal. Inspired by their work, our MPC architecture also makes use of subgoals to inform the optimization process but additionally adapts the pool of candidate trajectories to the context by propagating forward interaction-aware policies.


\subsection{Formalizing Cooperation in Crowd Navigation}

Acknowledging the prevalent role of cooperation~\citep{Wolfinger95} in crowd navigation, recent work incorporates the expectation of human cooperation and rationality into the design of frameworks for prediction and control~\citep{core-challenges2021}. For instance, \citet{trautmanijrr} developed a Gaussian-process-based human motion prediction model that estimates future human trajectories under the assumption of cooperative and goal-directed human behavior. Recently, \citet{SunM-RSS-21} formalized multiagent cooperation as a superposition of agents' probabilistic motion preferences, and developed a sampling-based, cooperation-aware motion planner. \citet{chaocao} developed a graph-search-based approach that plans robot motion by discovering channels of safe passages within a crowd. In past work, we formalized cooperation as inference over multiagent motion primitives represented as topological symbols~\citep{mavrogiannis-braids} and trajectory sets~\citep{mavrogiannis-hamiltonians}. This work set the foundation for Social Momentum (SM)~\citep{social-momentum-thri}, a real-time, reactive controller that generates robot motion by maximizing the magnitudes of the pairwise momenta defined between the robot and humans. This has the effect of producing actions that exaggerate over a passing side while moving the robot as far as possible. While this behavior is legible and positively perceived, it can be undesirable in more densely crowded environments with less space for maneuverability. 

In this paper instead, we formalize a topological notion of \emph{passing} based on a pairwise winding number~\citep{Berger2001invariants}, which enables the robot to monitor and expedite the progress of passing an agent. Topological representations offer interpretability and decomposition of challenging problems in robotics~\citep{Bhattacharya-RSS-11,knepper12,pokorny16,shkurti17,orthey21}. While prior work has also used winding numbers to capture multimodality in multiagent navigation, existing approaches do so in a binary sense to predict passing side preferences~\citep{kretzschmar_ijrr16,roh2020corl}, essentially discarding the notion of passing progress provided by the absolute value of the winding number. Instead, our approach is built around this notion of passing progress: out of a pool of candidate robot trajectories, our cost selects the one that contributes the most passing progress without bias over a passing side. Assuming goal-directed human behavior around the robot (humans moving from start to goal), minimizing our cost also has the effect of preserving a passing side. Our formalism is decoupled from the constraint of collision avoidance which is instead handled through a cost based on an explicit approximation of personal space~\citep{kirby_thesis}. Finally, unlike SM~\citep{social-momentum-thri}, which is essentially a reactive, rule-based controller with a one-step horizon, our framework incorporates arbitrary horizons through a more general MPC formulation yielding smoother performance.

\section{Preliminaries}\label{sec:preliminaries}

In this section, we formalize our domain of interest, and introduce the technical background of our approach.

\subsection{Problem Statement}

Consider a workspace $\mathcal{W}\subseteq\mathbb{R}^2$ where a robot navigates among $n$ dynamic agents. Denote by $s_k$ the state of the robot and by $s_k^i\in \mathcal{W}$, $i \in\mathcal{N} = \{1, \dots, n\}$, the state of agent $i$ at timestep $k$. The robot is navigating from a state $s_{0}$ towards a destination $s_{T}$ by executing controls $u$ from a space of controls $\mathcal{U}$, subject to its dynamics model $s_{k+1} = g(s_k,u_k)$. Agent $i\in\mathcal{N}$ is navigating from $s_{0}^i$ towards a destination $s_{T}^i$ by executing controls $u^i$ from a space of controls $\mathcal{U}^i$, striving for safety and efficiency. The robot is not aware of agent $i$'s destination $s^{i}_T$ or policy. However, we assume that the robot is perfectly observing the complete world state $(s_k, s_k^{1:n})$, where $s_k^{1:n} = (s^1_k,\dots, s^n_k)$. Our goal is to design a robot policy that enables the robot to navigate from $s_0$ to $s_T$ safely and efficiently.

\subsection{Model Predictive Control for Crowd Navigation}\label{sec:mpc}


We employ a discrete MPC formulation that optimizes a robot control trajectory $\boldsymbol{u}$ from a set $\mathbf{\mathcal{U}}$ of candidate trajectories of finite length $N$ with respect to a cost $\mathcal{J}$:

\begin{equation}
\begin{split}
\boldsymbol{u}^{*} = \arg & \min_{\boldsymbol{u}\in\boldsymbol{\mathcal{U}}} \mathcal{J}(\boldsymbol{s}, \boldsymbol{s}^{1:n})\\
    s.t.\: & s_{k+1} = g(s_k, u_k)\\
           & s^i_{k+1} = f(s_{k-h:k}, \boldsymbol{s}^{1:n}_{k-h:k}),\: i\in\mathcal{N}
   \label{eq:mpc}
\end{split}\mbox{,}
\end{equation}
where we denote by $\boldsymbol{s} = s_{1:N}$ the robot trajectory, by $\boldsymbol{s}^{1:n} = (\boldsymbol{s}^1,\dots, \boldsymbol{s}^n)$ the trajectories of agents $1,\dots, n$ lying in front of the robot, and by $\boldsymbol{u} = u_{0:N-1}$ a robot control trajectory, drawn from a set of short-horizon ($N$) trajectories $\boldsymbol{\mathcal{U}}$. Finally, we denote by $f$ a state transition model for agent $i$ that takes as input the system state history up to $h$ timesteps in the past. 


\subsection{V-MPC: Balancing Safety and Efficiency}\label{sec:vmpc}

Many recent crowd navigation controllers follow objectives related to safety and efficiency~\citep{Everett18_IROS,liu2020decentralized,roh2020corl,nishimura2020,SunM-RSS-21,wang2021corl}. We defined a baseline \emph{Vanilla} MPC (V-MPC) encoding such specifications through a cost $\mathcal{J}_{\nu}$:
\begin{equation}
    \mathcal{J}_{\nu}(\boldsymbol{s}, \boldsymbol{s}^{1:n}) = a_g \mathcal{J}_g(\boldsymbol{s}) + a_d \mathcal{J}_d(\boldsymbol{s}, \boldsymbol{s}^{1:n})\label{eq:vmpc-cost}\mbox{.}
\end{equation}
The term 
\begin{equation}
    \mathcal{J}_{g} (\boldsymbol{s}) = \sum_{k = 0}^{N-1} (s_{k+1}-s_T)^\intercal Q_g (s_{k+1}-s_T)\mbox{,}
\end{equation}
is a goal-tracking cost penalizing trajectories taking the robot further from its goal, where $Q_g$ is a weight matrix. The term
\begin{equation}
    J_d(\boldsymbol{s}, \boldsymbol{s}^{1:n}) = \sum_{k=0}^{N-1}\sum_{i=1}^n A_d^2(s_{k+1}, s_{k+1}^i)\mbox{,}
\end{equation}
is a cost penalizing violations to agents' personal space~\citep{hall-1966} through the \emph{Asymmetric Gaussian Integral Function} $A_d$ of \citet{kirby_thesis}. The weights $a_g, a_d$ encode the relative importance of cost terms. We approximate agents' state transition in~\eqref{eq:mpc} through a constant-velocity (CV) motion model $s^i_{k+1} = f(s^i_{k-1:k})$, propagating the velocity of agent $i$ one timestep $dt$ into the future, ignoring any interactions with other agents. 


\section{Topology-Informed Navigation\label{sec:topology}}

We introduce a mathematical representation that enables a robot to monitor the process of \emph{passing} other agents. Based on this representation, we design a cost function that motivates robot actions contributing more passing progress.


\begin{figure}
    \centering
         \begin{subfigure}{\linewidth}
     \centering
         \includegraphics[width = \linewidth]{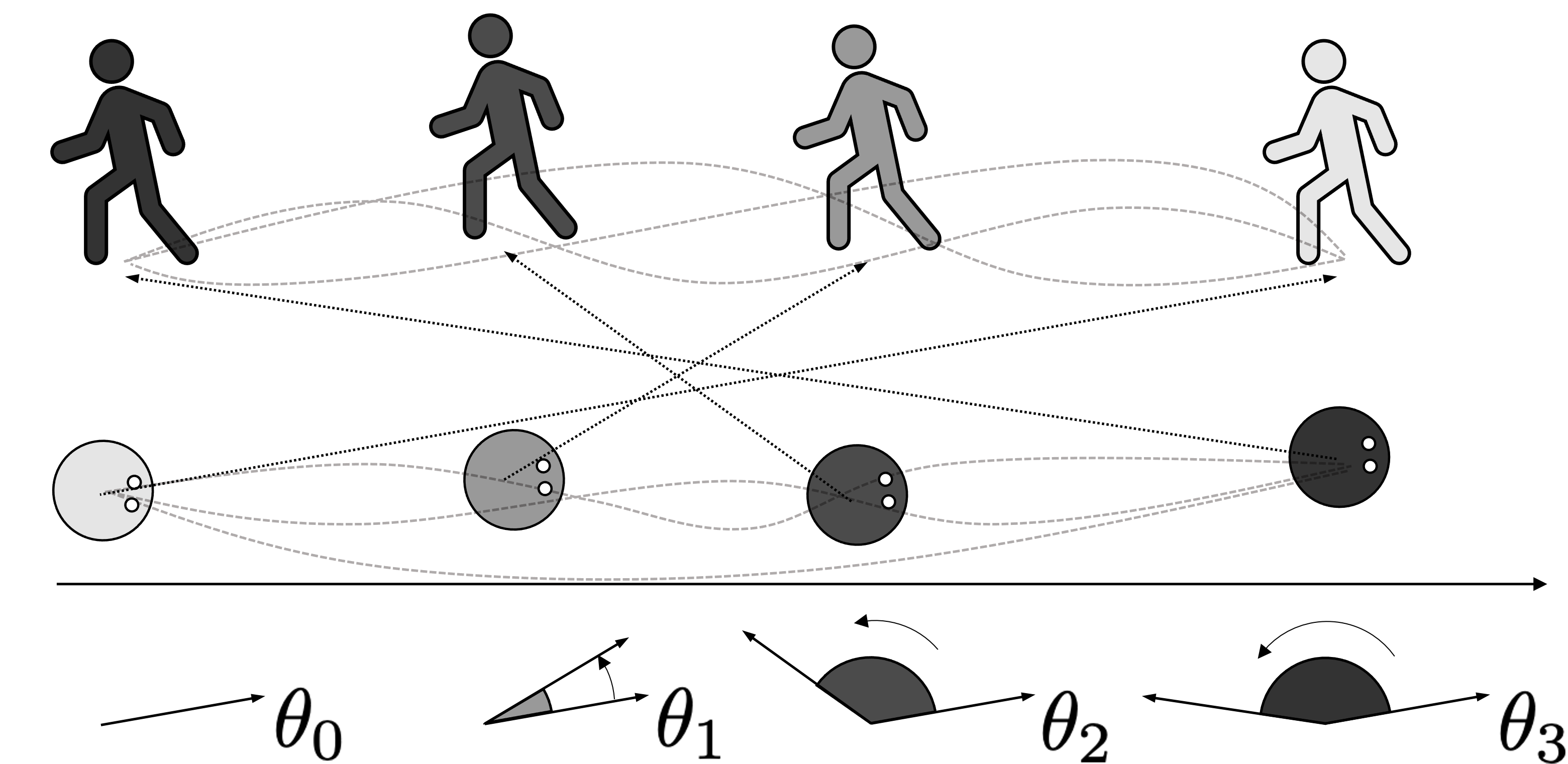}
         \caption{Passing from the right yields $\theta>0$.} 
         \label{fig:neglambda}
     \end{subfigure}
    \begin{subfigure}{\linewidth}
     \centering
         \includegraphics[width = \linewidth]{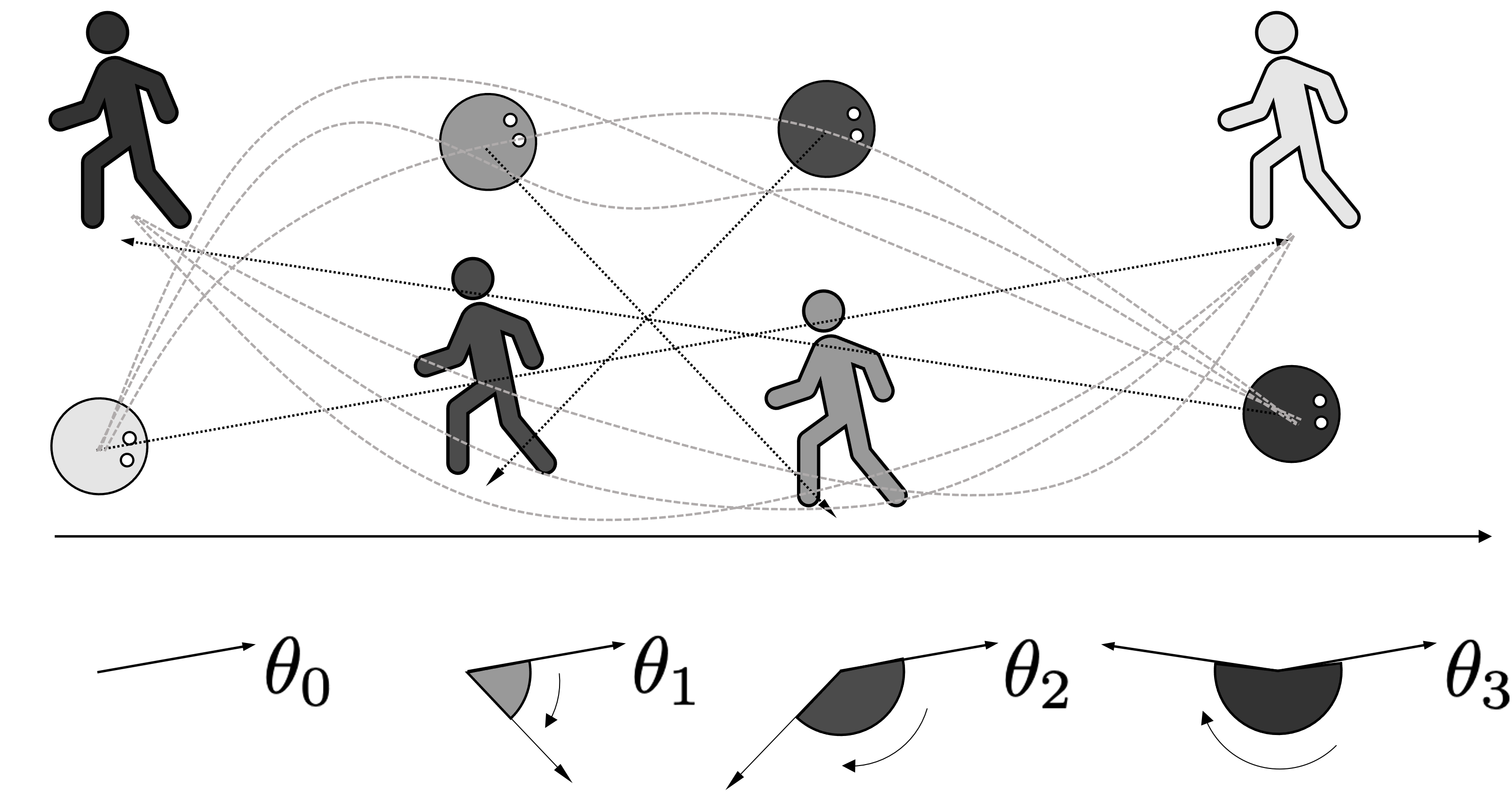}
         \caption{Passing from the left yields $\theta<0$.} 
         \label{fig:poslambda}
     \end{subfigure}
    \caption{Two different executions of the same scenario. As agents navigate (color intensity denotes timing), the vector connecting their positions rotates from $\theta_0$ to $\theta_3$. When they pass each other on the right, this rotation is counterclockwise (\subref{fig:neglambda}), whereas when they pass each other on the left (\subref{fig:poslambda}), it is clockwise. The winding number integral $\lambda = (\theta_3-\theta_0)/2\pi$ is a topological invariant, independent of the specific paths that agents take, describing the signed number of times that agents pass each other.}
    \label{fig:invariance}
\end{figure}


\subsection{Formalizing Passing as Pairwise Rotation}\label{sec:windingnumbers}

We denote by $\mathrm{x}_{k}^{i}\in\mathbb{R}^2$ the position of agent $i$ relative to the robot at timestep $k$, and by $\theta^{i}_k = \angle\mathrm{x}_{k}^{i}$ the angle of that vector with respect to a fixed global frame. From time $k$ to $k + 1$, agents' displacement from $\mathrm{x}_{k}^{i}$ to $\mathrm{x}_{k+1}^{i}$ results in a rotation $\Delta \theta^{i}_{k+1} = \theta^{i}_{k+1} - \theta^{i}_k$. We denote by $\boldsymbol{s} = s_{1:N} = (s_1, \dots, s_N)$ the robot's trajectory and by $\boldsymbol{s}^{i} = s_{1:N}^i = (s_1^i, \dots, s_N^i)$ the trajectory of agent $i\in\mathcal{N}$. The quantity
\begin{equation}
\lambda^{i}(\boldsymbol{s},\boldsymbol{s}^i) = \frac{1}{2\pi} \sum_{k = 0}^{N-1} \Delta \theta^{i}_{k+1}\mbox{,}\label{eq:windingnumber}
\end{equation}
is a \emph{winding number}~\citep{Berger2001invariants} representing the signed number of times that the robot and agent $i$ revolved around each other over a horizon of $N$ timesteps. In the crowd navigation domains of our focus, where agents exhibit goal-directed, cooperative behavior, the winding number captures the progress (magnitude), and side (sign) of \emph{passing} each other (see~\figref{fig:invariance}).
A right-side passing induces a counterclockwise rotation yielding a positive winding number ($\lambda^{i} > 0$), whereas a left-side passing yields a clockwise rotation and thus a negative winding number ($\lambda^{i} < 0$).

\begin{figure}
     \centering
         \includegraphics[width = \linewidth]{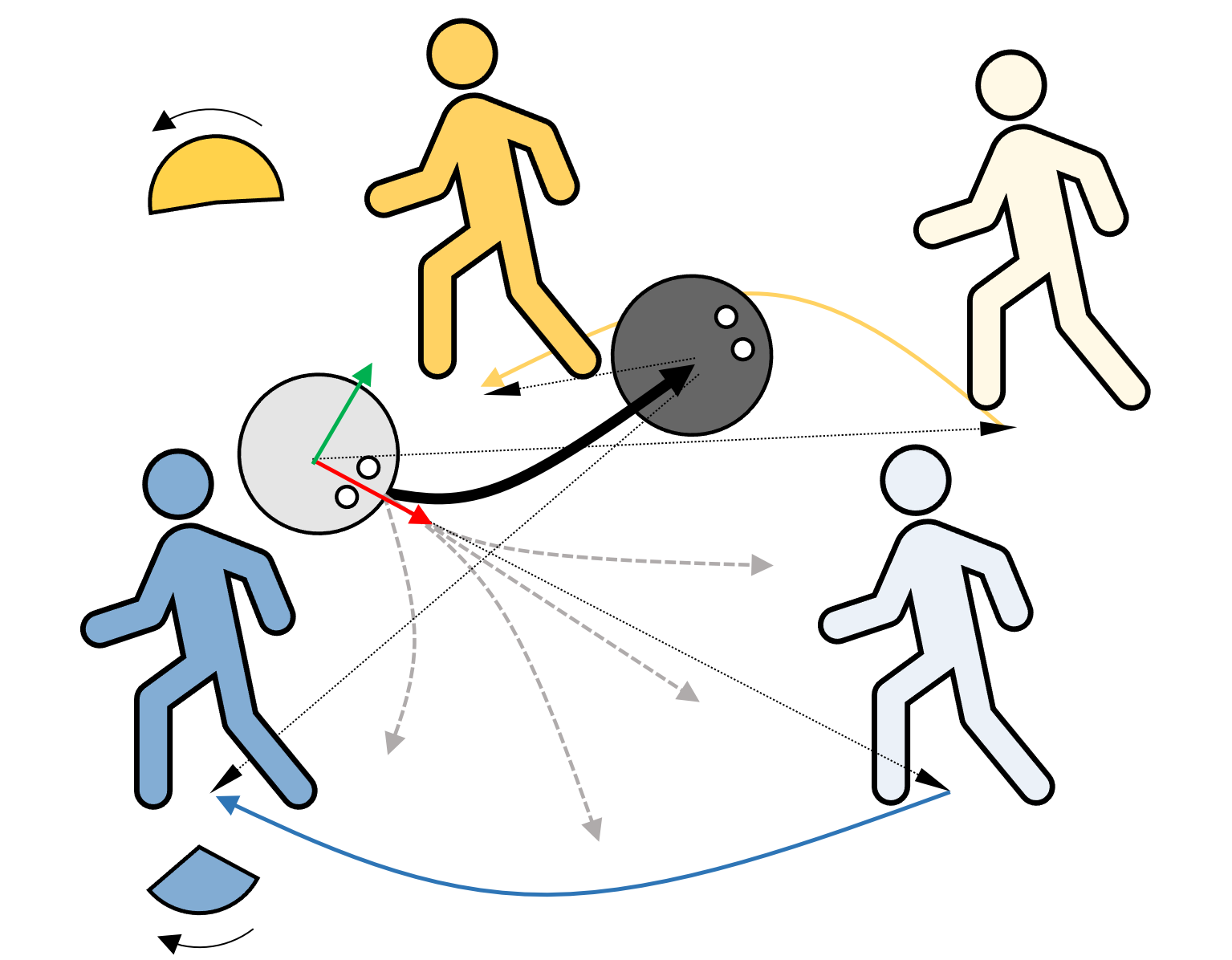}
         \label{fig:threeagents}
    \caption{Effect of minimizing the passing cost. The robot predicts how the blue and yellow agents will move and selects an action (black arrow) that maximizes the average passing progress over the control horizon (color intensity indicates timing). The circular sectors denote the estimated progress to be achieved for each agent. The robot ignores agents lying behind it, i.e., on the negatives of its body $x$-axis (red color).}
    \label{fig:example}
\end{figure}

\subsection{A Cost Functional that Expedites Passing}

Our key insight is that by monitoring and contributing to the values of $\lambda^i$, $i\in\mathcal{N}$, through its own actions, the robot may anticipate, influence, and proactively adapt to the passing sides of humans without relying on high-accuracy models of human motion prediction. To engineer robot motion that expedites the process of passing between the robot and other agents, we design the passing cost
\begin{equation}
    \mathcal{J}_p(\boldsymbol{s}, \boldsymbol{s}^{1:n}) = -\frac{1}{n}\sum_{i=1}^{n} \lambda^i(\boldsymbol{s},\boldsymbol{s}^i)^2\mbox{.}\label{eq:topologycost}
\end{equation}
We assume that the future trajectories of human agents $\boldsymbol{s}^{1:n}$ are estimated through a transition model $f$ as shown in~\eqref{eq:mpc}, and provided as input in $\mathcal{J}_p$. Given $\boldsymbol{s}^{1:n}$, the minimization of $\mathcal{J}_p$ over $\boldsymbol{s}$ yields a robot trajectory $\boldsymbol{s}^*$ that maximizes (on average) the winding numbers between the robot and human agents. This trajectory enables the robot to maneuver itself in a way that aligns with the estimated passing intentions of others. This corresponds to robot motion that avoids switching between passing sides as it approaches a human, expediting and facilitating collision avoidance. \figref{fig:example} gives intuition about the effect of minimizing the proposed cost. Note that the cost of eq.~\eqref{eq:topologycost} excludes non-reactive agents lying behind the robot.

\subsection{Topology-Informed MPC}

The topology-informed MPC (T-MPC) is an extension of V-MPC~\eqref{eq:vmpc-cost} that incorporates the functional of \eqref{eq:topologycost}:
\begin{equation}
    \mathcal{J}(\boldsymbol{s}, \boldsymbol{s}^{1:n}) = \mathcal{J}_{\nu} + a_p \mathcal{J}_p(\boldsymbol{s}, \boldsymbol{s}^{1:n})\label{eq:tmpc}\mbox{,}
\end{equation}
where and $a_p$ is a weight of relative significance. T-MPC strives to expedite passing between the robot and other agents while respecting their personal space and approaching its destination. In conjunction, this formulation is designed to motivate goal-oriented, anticipatory collision avoidance.

\textbf{Remark}: As discussed in Sec.~\ref{sec:mpc}, our core MPC formulation is an optimization over a discrete set of short-horizon motion primitives. These primitives are scored with respect to a cost computed by taking into account only agents lying in front of the robot (see~\figref{fig:example}). Thus, the minimization of the $\mathcal{J}_p$ cost in~\eqref{eq:tmpc} maximizes the average progress with respect to agents lying in front of the robot. Through subsequent control cycles, the minimization of $\mathcal{J}_p$ contributes consistent progress to the prevailing passing side; switching to a different side gets progressively more penalized since it would correspond to a lower winding number as two agents approach each other. Once the robot passes a human agent, it stops considering them in the computation of costs, thus avoiding possible undesired effects like circling around them.

\section{Evaluation}\label{sec:evaluation}

We evaluated our controller through simulated and real-world experiments involving robot navigation in crowded scenes. We used Honda's experimental ballbot~\citep{pathbot} (see~\figref{fig:introfig}) and its own low-level velocity controller~\citep{yamane2019}. A ballbot~\citep{oneisenough} is a mobile robot with a mechanical body dynamically balancing on top of an omnidirectional spherical wheel. The ballbot's dynamics produces readable~\citep{Lo19} and safe behavior~\citep{oneisenough}, making it amenable for operation in crowded spaces.

\subsection{Baselines}


We evaluate our MPC against two baselines from the literature:

\textbf{CADRL}~\citep{Everett18_IROS} is a deep RL framework for multiagent collision avoidance. We adapt the original formulation to handle ballbot dynamics by augmenting: a) its state $\mathbf{s}$ to include the ballbot's inclination $\phi$, and the average distance to the robot's goal over the past $t'$ timesteps, $\bar{d}_g$; b) its reward function to include an inclination penalty term

\begin{equation}
R_{lean}(\mathbf{s}) =
    \begin{cases}
    -1, & \textnormal{if}\: \phi > \phi_{max}\\
    -0.1\frac{\phi}{\phi{max}} , & \textnormal{otherwise}\mbox{,}
    \end{cases}
\label{rlean}
\end{equation}
where $\phi_{max}$ is an inclination threshold (set to $0.25$, if $\phi > \phi_{max}$ the episode terminates), and a reward of progress to goal $R_{prog}(\mathbf{s}) = 0.1 \cdot (\bar{d}_g - d_{g}(\mathbf{s}))$, where $d_g(s)$ is the robot's distance to goal. We included the ballbot dynamics as part of the state transition model. Initializing from the implementation of~\citep{Everett18_IROS}, we trained the model in two stages, first with 2-4 agents and then with 2-10 agents.

\begin{figure}
    \centering
    \begin{subfigure}{.31\linewidth}
    \centering
        \includegraphics[width = \linewidth]{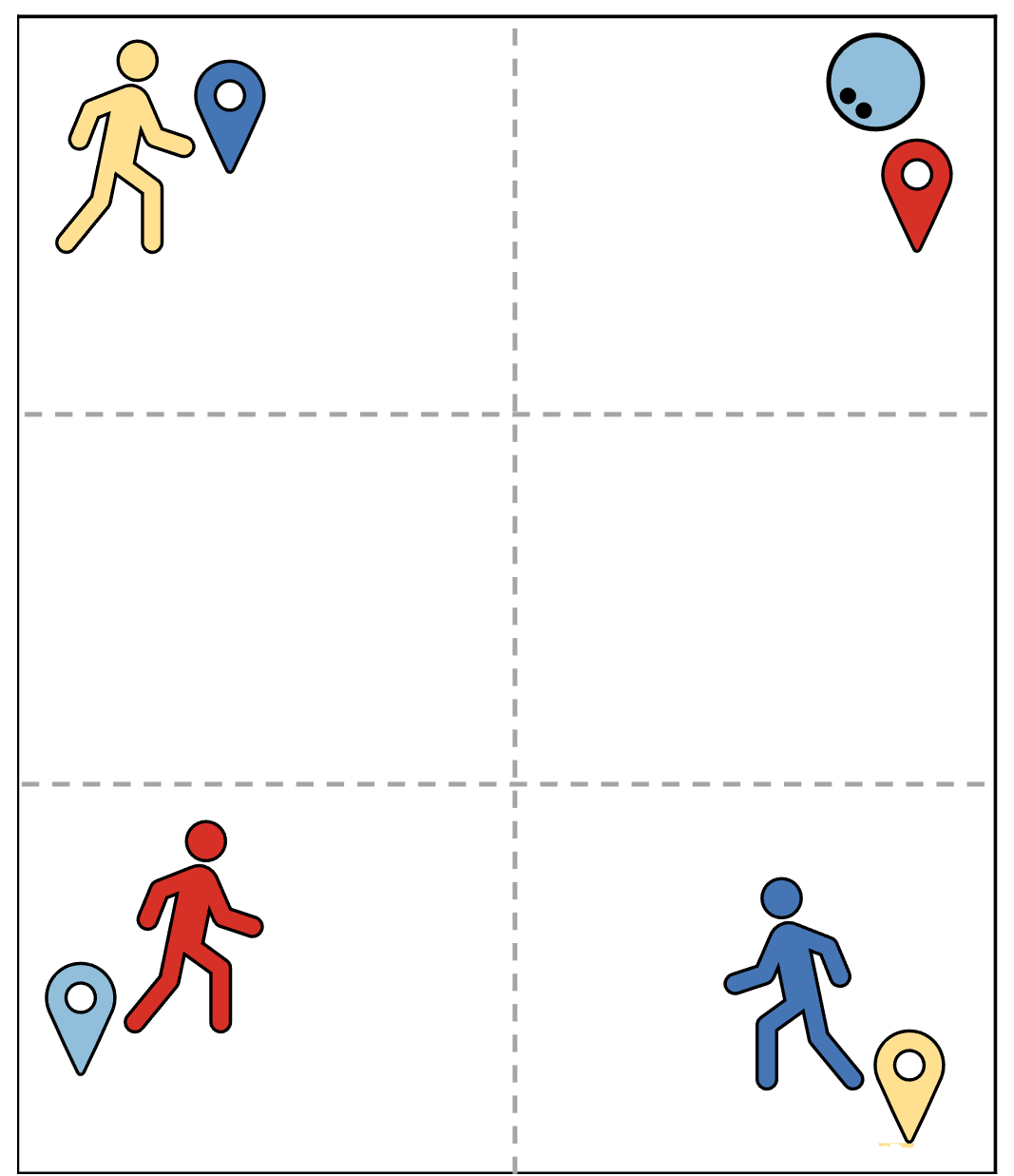}
        \caption{Three humans.}
        \label{fig:3agents}
    \end{subfigure}
    \begin{subfigure}{.31\linewidth}
     \centering
         \includegraphics[width = \linewidth]{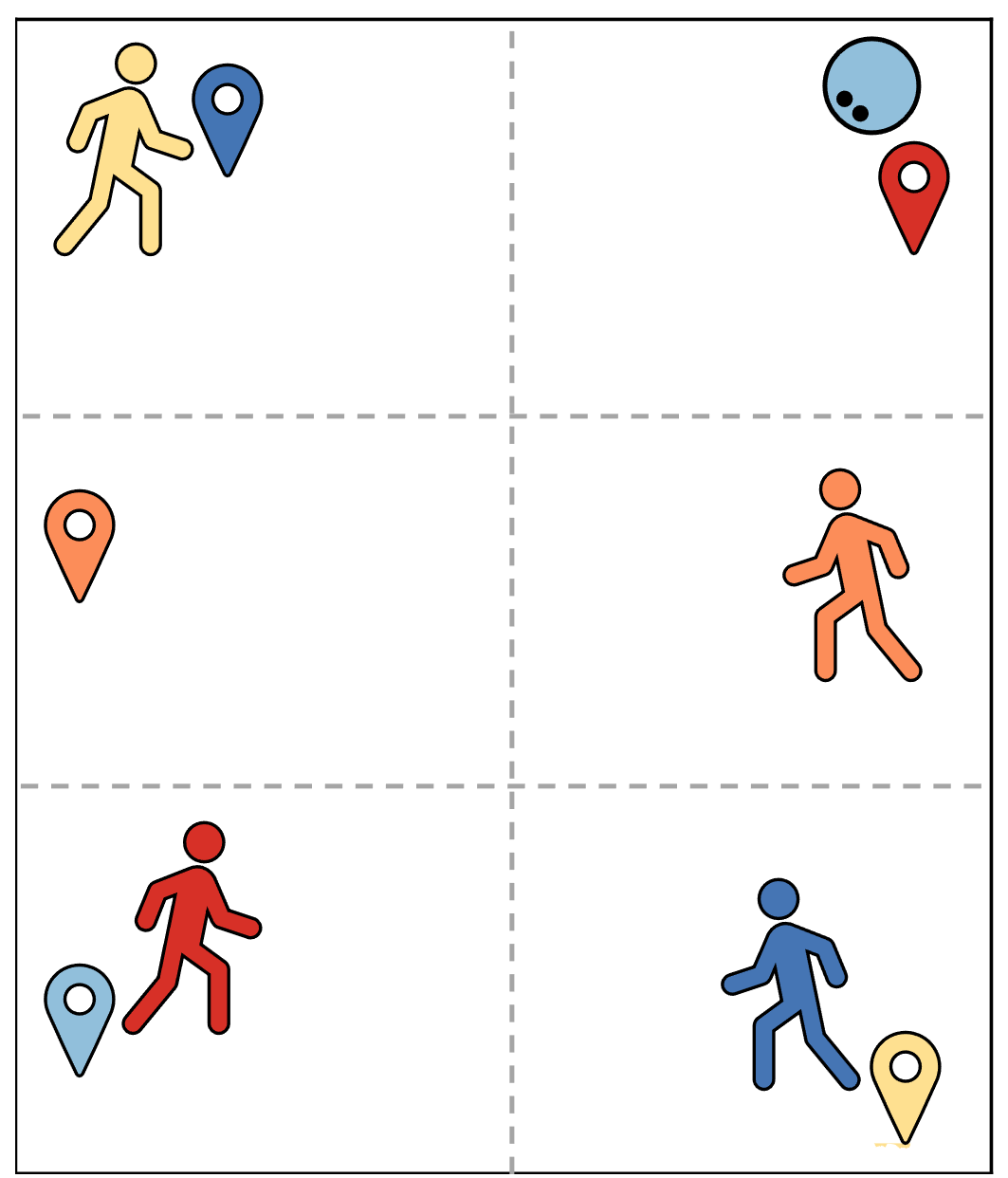}
         \caption{Four humans.} 
         \label{fig:4agents}
     \end{subfigure}
    \begin{subfigure}{.31\linewidth}
    \centering
        \includegraphics[width = \linewidth]{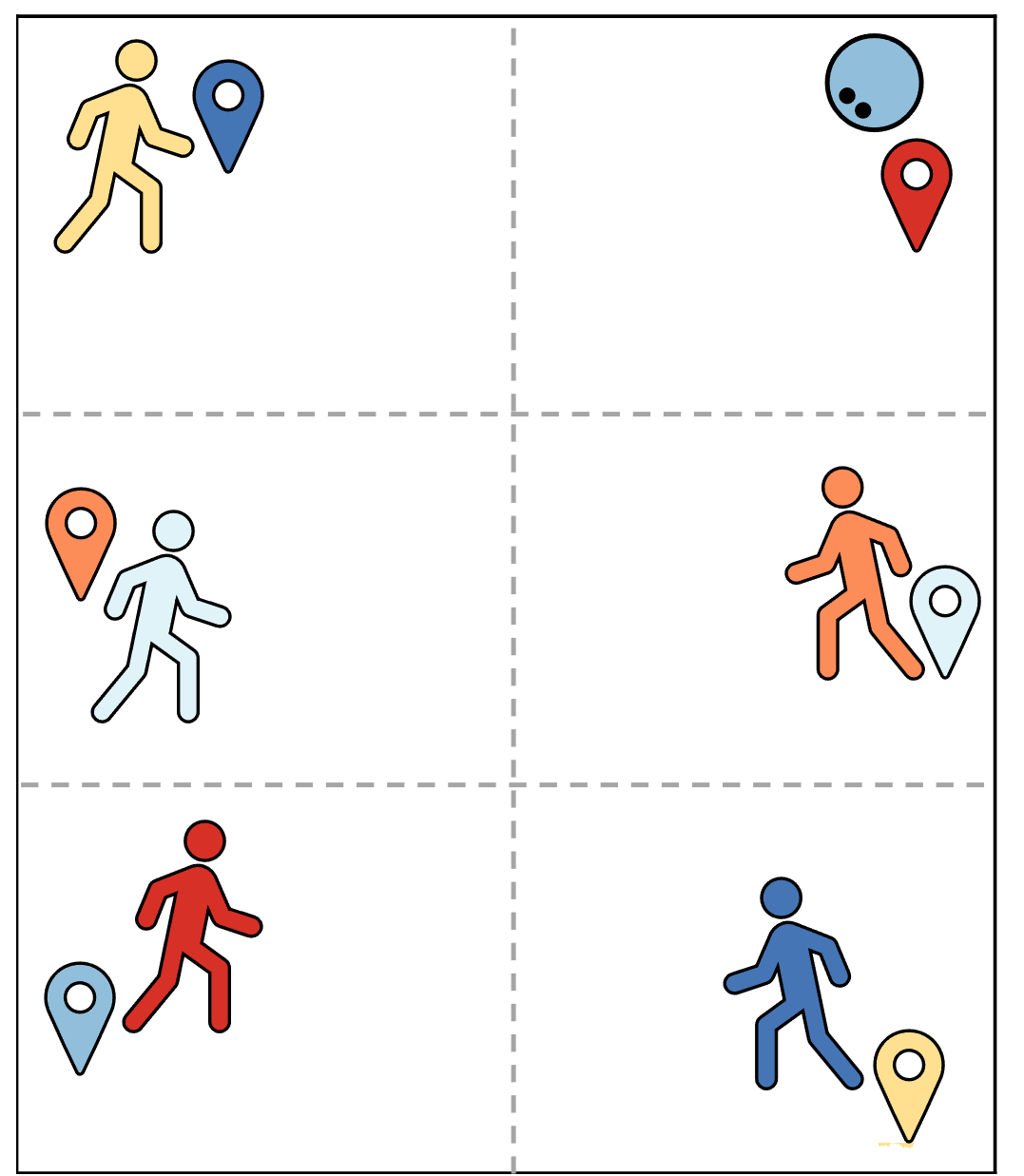}
        \caption{Five humans.} \label{fig:5agents}
    \end{subfigure}
\caption{The scenarios used in our evaluation. The workspace is split in 6 zones. We draw agents in their starting zones and show their destination zones as landmarks of the same color. \label{fig:scenarios}}    
\end{figure}

\textbf{ORCA}~\citep{ORCA} is a crowd simulator that is often used in the evaluation of crowd navigation algorithms~\citep{Everett18_IROS,social-momentum-thri,chaocao,chen2019crowd,liu2020decentralized,liu2020decentralized,core-challenges2021}. We employ the ORCA configuration of~\citet{chen2019crowd}.




\subsection{MPC Implementation}\label{sec:mpc-implementation}
 
Motivated by the approach of~\citet{brito-ral21}, our MPC employs a set $\bm{\mathcal{U}}$ of rollouts (robot control trajectories) extracted by propagating a policy $\pi_{sim}$ towards $m=10$ subgoals, placed around the robot at fixed orientation intervals of $\frac{\pi}{5}$ and distance of $8m$ for 10 timesteps of size $dt=0.1s$. We experimented with three policies: CV, in which we propagate the robot's state forward assuming a constant preferred speed of $0.8m/s$ without collision avoidance considerations; ORCA~\citep{ORCA}; CADRL~\citep{Everett18_IROS}. For human motion prediction $f$ we used a CV approximation, propagating humans' velocities forward. We tuned all MPC variants through parameter sweeps for \emph{Safety} $ \mathcal{D}$ $(m)$, defined as the minimum distance between the robot and human agents throughout a trial, and \emph{Time to destination} $\mathcal{T}$ $(s)$ over 30 randomized trials for each scenario of~\figref{fig:scenarios}. The control loop closed with frequency $10Hz$.

\begin{table*}
\centering
\caption{Performance with respect to Safety ($\mathcal{D}$) and Time to destination ($\mathcal{T}$) per world. Each entry contains a mean and a standard deviation over 100 trials. The best performing controller per column is printed in bold. Green, orange, and blue entries are controllers outperformed by the best at significance level of $p<0.001$, $<0.01$, and $<0.05$ respectively (U test). 
\label{tab:results}
}
\resizebox{\linewidth}{!}{%
\addvbuffer[0pt 0pt]{
\begin{tabular}{l|cccccccccccc}
\toprule
\multicolumn{1}{l}{\textbf{Scenario}} &
  \multicolumn{4}{c}{Three Humans} &
  \multicolumn{4}{c}{Four Humans} &
  \multicolumn{4}{c}{Five Humans} \\ \midrule
\multicolumn{1}{l}{\textbf{Metric}} &
  \multicolumn{2}{c|}{$\mathcal{D} (m)$} &
  \multicolumn{2}{c|}{$\mathcal{T} (s)$} &
  \multicolumn{2}{c|}{$\mathcal{D} (m)$} &
  \multicolumn{2}{c|}{$\mathcal{T} (s)$} &
  \multicolumn{2}{c|}{$\mathcal{D} (m)$} &
  \multicolumn{2}{c}{$\mathcal{T} (s)$} \\ \midrule 
\multicolumn{1}{l}{\textbf{World}} &
  ORCA &
  \multicolumn{1}{c|}{CADRL} &
  ORCA &
  \multicolumn{1}{c|}{CADRL} &
  ORCA &
  \multicolumn{1}{c|}{CADRL} &
  ORCA &
  \multicolumn{1}{c|}{CADRL} &
  ORCA &
  \multicolumn{1}{c|}{CADRL} &
  ORCA &
  \multicolumn{1}{c}{CADRL} \\ \midrule 
\multicolumn{1}{l|}{ORCA} &
  \textcolor{MYgreen}{0.67 $\pm$ 0.14} &
  \multicolumn{1}{c|}{\textcolor{MYgreen}{0.93 $\pm$ 0.26}} &
  \textbf{9.33 $\pm$ 0.44} &
  \multicolumn{1}{c|}{{10.55 $\pm$ 1.13}} &
  \textcolor{MYgreen}{0.64 $\pm$ 0.12} &
  \multicolumn{1}{c|}{\textcolor{MYgreen}{0.80 $\pm$ 0.26}} &
  \textbf{9.70 $\pm$ 1.33} &
  \multicolumn{1}{c|}{{11.01 $\pm$ 1.96}} &
  \textcolor{MYgreen}{0.61 $\pm$ 0.05} &
  \multicolumn{1}{c|}{\textcolor{MYgreen}{0.69 $\pm$ 0.18}} &
  \textbf{10.08 $\pm$ 1.57} &
  \multicolumn{1}{c}{\textcolor{MYdarkblue}{11.37 $\pm$ 2.32}} \\
\multicolumn{1}{l|}{CADRL} &
  \textcolor{MYgreen}{0.68 $\pm$ 0.19} &
  \multicolumn{1}{c|}{\textcolor{MYgreen}{0.97 $\pm$ 0.31}} &
    {9.82 $\pm$ 1.40} &
  \multicolumn{1}{c|}{\textbf{10.50 $\pm$ 1.60}} &
  \textcolor{MYgreen}{0.63 $\pm$ 0.11} &
  \multicolumn{1}{c|}{\textcolor{MYgreen}{0.91 $\pm$ 0.24}} &
  \textcolor{MYgreen}{10.34 $\pm$ 1.65} &
  \multicolumn{1}{c|}{\textbf{10.62 $\pm$ 1.49}} &
  \textcolor{MYgreen}{0.6 $\pm$ 0.06} &
  \multicolumn{1}{c|}{\textcolor{MYgreen}{0.72 $\pm$ 0.18}} &
  12.32 $\pm$ 2.85 &
  \multicolumn{1}{c}{\textcolor{MYgreen}{12.86 $\pm$ 3.42}} \\ \midrule
\multicolumn{1}{l|}{V-MPC-CV} &
  \textcolor{MYgreen}{0.64 $\pm$ 0.13} &
  \multicolumn{1}{c|}{\textcolor{MYgreen}{0.93 $\pm$ 0.29}} &
  \textcolor{MYgreen}{10.87 $\pm$ 1.11} &
  \multicolumn{1}{c|}{\textcolor{MYlightblue}{11.26 $\pm$ 1.37}} &
  \textcolor{MYgreen}{0.61 $\pm$ 0.10} &
  \multicolumn{1}{c|}{\textcolor{MYgreen}{0.80 $\pm$ 0.28}} &
  \textcolor{MYgreen}{10.77 $\pm$ 1.26} &
  \multicolumn{1}{c|}{\textcolor{MYlightblue}{11.90 $\pm$ 2.02}} &
  \textcolor{MYgreen}{0.59 $\pm$ 0.09} &
  \multicolumn{1}{c|}{\textcolor{MYgreen}{0.61 $\pm$ 0.17}} &
  \textcolor{MYgreen}{11.27 $\pm$ 2.14} &
  \multicolumn{1}{c}{\textcolor{MYdarkblue}{12.70 $\pm$ 3.17}} \\
\multicolumn{1}{l|}{V-MPC-ORCA} &
   0.74 $\pm$ 0.24 &
  \multicolumn{1}{c|}{\textcolor{MYdarkblue}{1.06 $\pm$ 0.29}} &
  \textcolor{MYgreen}{11.14 $\pm$ 1.45} &
  \multicolumn{1}{c|}{\textcolor{MYgreen}{12.14 $\pm$ 1.97}} &
  \textcolor{MYlightblue}{0.66 $\pm$ 0.14} &
  \multicolumn{1}{c|}{\textcolor{MYlightblue}{1.01 $\pm$ 0.26}} &
  \textcolor{MYgreen}{11.66 $\pm$ 1.70} &
  \multicolumn{1}{c|}{\textcolor{MYgreen}{12.01 $\pm$ 1.25}} &
  \textcolor{MYlightblue}{0.65 $\pm$ 0.12} &
  \multicolumn{1}{c|}{\textcolor{MYgreen}{0.77 $\pm$ 0.24}} &
  \textcolor{MYgreen}{12.17 $\pm$ 3.02} &
  \multicolumn{1}{c}{\textcolor{MYgreen}{{13.18 $\pm$ 3.12}}} \\
\multicolumn{1}{l|}{V-MPC-CADRL} &
  \textcolor{MYgreen}{0.64 $\pm$ 0.15} &
  \multicolumn{1}{c|}{\textcolor{MYgreen}{0.88 $\pm$ 0.33}} &
  \textcolor{MYgreen}{12.61 $\pm$ 3.22} &
  \multicolumn{1}{c|}{\textcolor{MYgreen}{12.29 $\pm$ 1.54}} &
  \textcolor{MYgreen}{0.65 $\pm$ 0.15} &
  \multicolumn{1}{c|}{\textcolor{MYgreen}{0.89 $\pm$ 0.36}} &
  \textcolor{MYgreen}{11.97 $\pm$ 2.63} &
  \multicolumn{1}{c|}{\textcolor{MYgreen}{12.36 $\pm$ 2.29}} &
  \textcolor{MYgreen}{0.61 $\pm$ 0.14} &
  \multicolumn{1}{c|}{\textcolor{MYgreen}{0.69 $\pm$ 0.25}} &
  \textcolor{MYgreen}{15.21 $\pm$ 4.10} &
  \multicolumn{1}{c}{\textcolor{MYgreen}{{13.68 $\pm$ 3.09}}} \\ \midrule
\multicolumn{1}{l|}{T-MPC-CV} &
  \textcolor{MYgreen}{0.67 $\pm$ 0.16} &
  \multicolumn{1}{c|}{1.08 $\pm$ 0.31} &
  \textcolor{MYgreen}{10.60 $\pm$ 1.16} &
  \multicolumn{1}{c|}{\textcolor{MYgreen}{11.15 $\pm$ 1.50}} &
  \textcolor{MYgreen}{0.64 $\pm$ 0.13} &
  \multicolumn{1}{c|}{1.09 $\pm$ 0.31} &
  \textcolor{MYgreen}{{10.44 $\pm$ 0.99}} &
  \multicolumn{1}{c|}{{10.68 $\pm$ 0.90}} &
  \textcolor{MYgreen}{0.63 $\pm$ 0.13} &
  \multicolumn{1}{c|}{\textcolor{MYdarkblue}{0.83 $\pm$ 0.25}} &
  \textcolor{MYgreen}{10.94 $\pm$ 1.49} &
  \multicolumn{1}{c}{\textbf{10.88 $\pm$ 1.14}} \\
\multicolumn{1}{l|}{T-MPC-ORCA} &
  \textbf{0.83 $\pm$ 0.34} &
  \multicolumn{1}{c|}{1.13 $\pm$ 0.31} &
  \textcolor{MYgreen}{11.50 $\pm$ 2.21} &
  \multicolumn{1}{c|}{\textcolor{MYgreen}{12.15 $\pm$ 1.83}} &
  \textbf{0.78 $\pm$ 0.23} &
  \multicolumn{1}{c|}{1.11 $\pm$ 0.32} &
  \textcolor{MYgreen}{11.11 $\pm$ 1.98} &
  \multicolumn{1}{c|}{\textcolor{MYgreen}{11.79 $\pm$ 2.09}} &
  \textbf{0.70 $\pm$ 0.15} &
  \multicolumn{1}{c|}{0.87 $\pm$ 0.26} &
  \textcolor{MYgreen}{{12.66 $\pm$ 3.52}} &
  \multicolumn{1}{c}{\textcolor{MYgreen}{{13.08 $\pm$ 3.28}}} \\
\multicolumn{1}{l|}{T-MPC-CADRL} &
  0.77 $\pm$ 0.26 &
  \multicolumn{1}{c|}{\textbf{1.17 $\pm$ 0.38}} &
  \textcolor{MYgreen}{11.15 $\pm$ 2.05} &
  \multicolumn{1}{c|}{\textcolor{MYgreen}{11.73 $\pm$ 1.57}} &
  0.75 $\pm$ 0.25 &
  \multicolumn{1}{c|}{\textbf{1.14 $\pm$ 0.38}} &
  \textcolor{MYgreen}{11.09 $\pm$ 1.98} &
  \multicolumn{1}{c|}{\textcolor{MYgreen}{11.13 $\pm$ 1.46}} &
  \textcolor{MYgreen}{0.66 $\pm$ 0.17} &
  \multicolumn{1}{c|}{\textbf{0.87 $\pm$ 0.28}} &
  \textcolor{MYgreen}{{14.19 $\pm$ 3.08}} & \textcolor{MYgreen}{{12.65 $\pm$ 2.61}}\\ \bottomrule
\end{tabular}
}
}
\end{table*}





\subsection{Experiment Design}\label{sec:simulations}

We conducted our evaluation in a workspace of area $3.6\times 4.5 m^2$. We partitioned the workspace into six zones of area $1.8\times 1.5m^2$, and defined three scenarios involving 3, 4, and 5 humans, as shown in~\figref{fig:scenarios}. In each scenario, humans move between start and goal zones, selected to give rise to challenging human-robot encounters. Across scenarios, the robot's start and goal coordinates are fixed at $(0,0)$ and $(3.6, 4.5)$, respectively. The preferred speed for the robot is set to $0.8m/s$ which was found to be a natural human walking speed in pilot trials. We evaluated robot performance with respect to \emph{Safety} and \emph{Time to destination}.

Overall, we anticipated that MPC performance would improve when using rollouts from reactive policies like CADRL and ORCA since the robot would have an informed way of reacting to human motion estimates compared to the CV baseline. Further, we anticipated that T-MPC would outperform all baselines in terms of \emph{Safety} since the passing cost would enable the robot to make consistent passing progress, avoiding approaching humans by switching passing sides. We formalized these insights into the following hypotheses:
\begin{itemize}

    \item [\textbf{H1}:] MPC controllers with ORCA or CADRL rollouts outperform controllers with CV rollouts in terms of Safety across all scenarios and worlds.

    \item [\textbf{H2:}] T-MPC outperforms V-MPC with identical rollouts across all scenarios (3, 4, 5 agents) and worlds (ORCA, CADRL) in terms of Safety.
    
    
    \item [\textbf{H3:}] T-MPC outperforms both CADRL and ORCA across all scenarios and worlds in terms of Safety.
\end{itemize}

\textbf{Simulations}: We instantiated all scenarios from~\figref{fig:scenarios} in two Gazebo~\citep{Gazebo} worlds: one in which humans are simulated as ORCA agents~\citep{ORCA}, and one as CADRL~\citep{Everett18_IROS} agents. For each scenario, we generated 100 trials by sampling start and goal coordinates for agents, uniformly at random from their assigned zones. We executed each batch of trials with the following policies: 1) ORCA; 2) CADRL; 3) V-MPC-CV; 4) V-MPC-ORCA; 5) V-MPC-CADRL; 6) T-MPC-CV; 7) T-MPC-ORCA; 8) T-MPC-CADRL. Across all simulations, humans are represented as spheres of $0.3m$ radius whereas the robot's body is modeled as a cylinder of radius 0.2m.

\textbf{Real-world experiments}: We instantiated the scenario of~\figref{fig:3agents} in a lab workspace, deploying Honda's ballbot~\citep{pathbot} to navigate next to three members of our research team (see~\figref{fig:introfig}). We ran 60 trials of the scenario in which the robot navigated with three different policies (20 trials per policy): a) CV in which the robot drives to the goal with constant velocity without avoiding collisions; b) CADRL; c) T-MPC-CADRL. The users were told to ``navigate towards the destinations designated by the scenario with normal walking speed and to treat the robot as a walking human''. The users and the robot wore hats with reflective markers that enabled high-accuracy localization through an Optitrack motion-capture system of twelve overhead cameras operating at $120Hz$.

\begin{figure}
    \centering
    \begin{subfigure}{\linewidth}
     \centering
         \includegraphics[width = \linewidth]{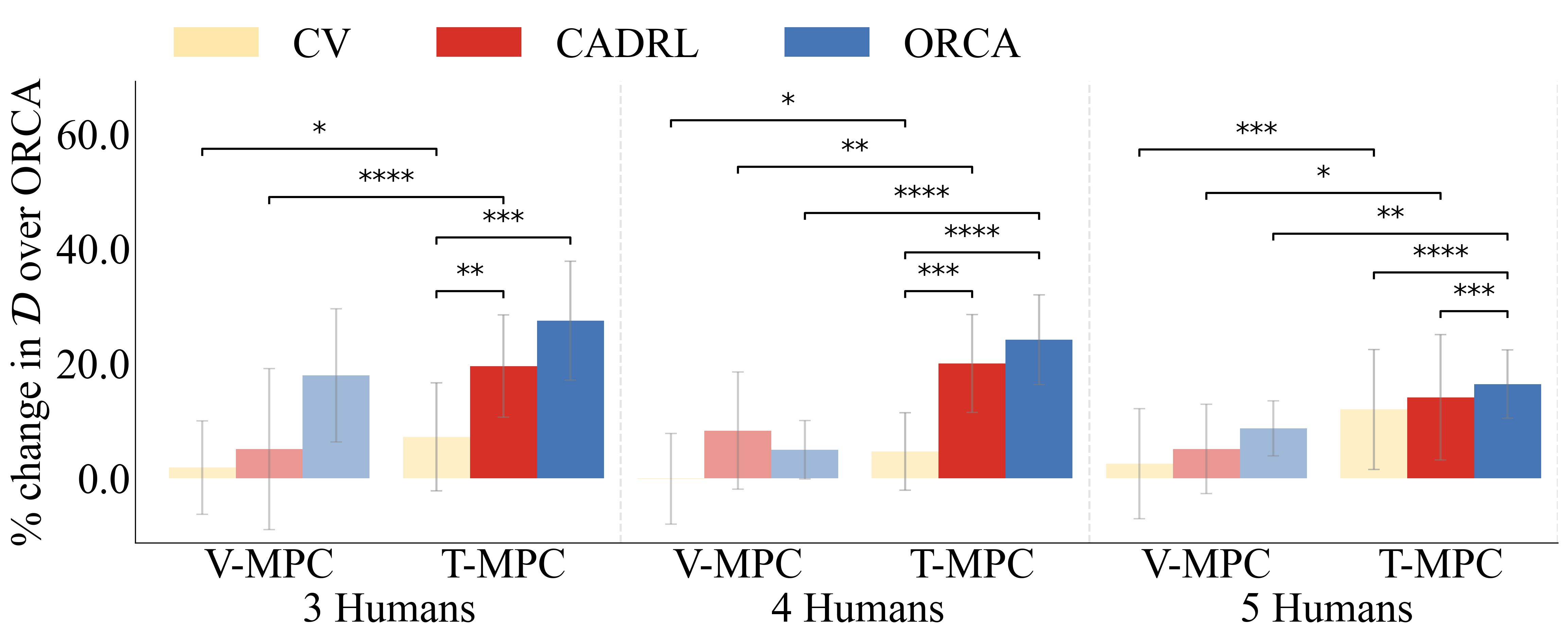}
         \caption{ORCA world.} 
         \label{fig:safety-orca}
     \end{subfigure}\\
    \begin{subfigure}{\linewidth}
    \centering
        \includegraphics[width = \linewidth]{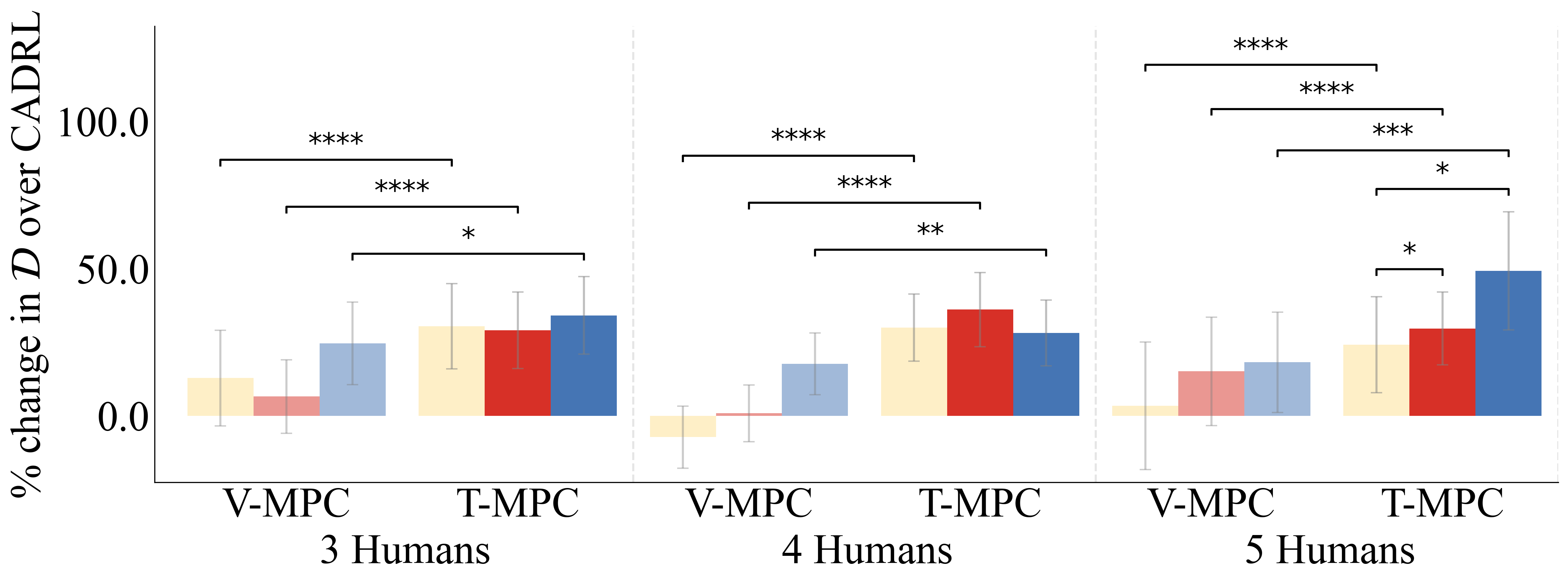}
        \caption{CADRL world.} \label{fig:safety-cadrl}
    \end{subfigure}
\caption{Comparison of MPC variants in terms of $\%$ change in Safety over the baselines of the two worlds. Error bars indicate $95\%$ confidence intervals. Stars indicate significance levels ($p<0.05$, $p<0.01$, $p<0.001$) according to a U test.\label{fig:safety-plots}}    
\end{figure}

\subsection{Results}

\begin{figure}
    \centering
    \begin{subfigure}{.48\linewidth}
    \centering
        \includegraphics[width=\linewidth]{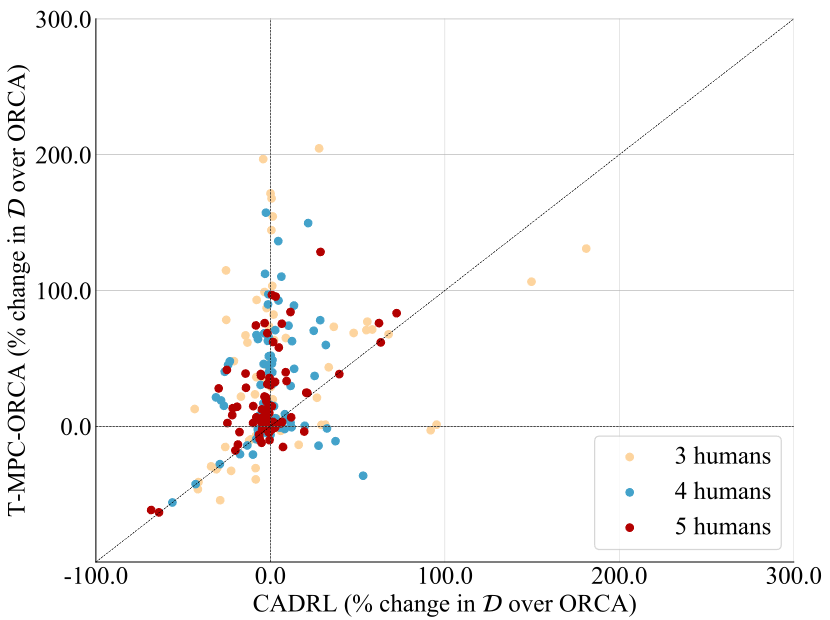}
        \caption{ORCA world.\label{fig:scatter-orcaworld}}
    \end{subfigure}
    \begin{subfigure}{.48\linewidth}
    \centering
        \includegraphics[width=\linewidth]{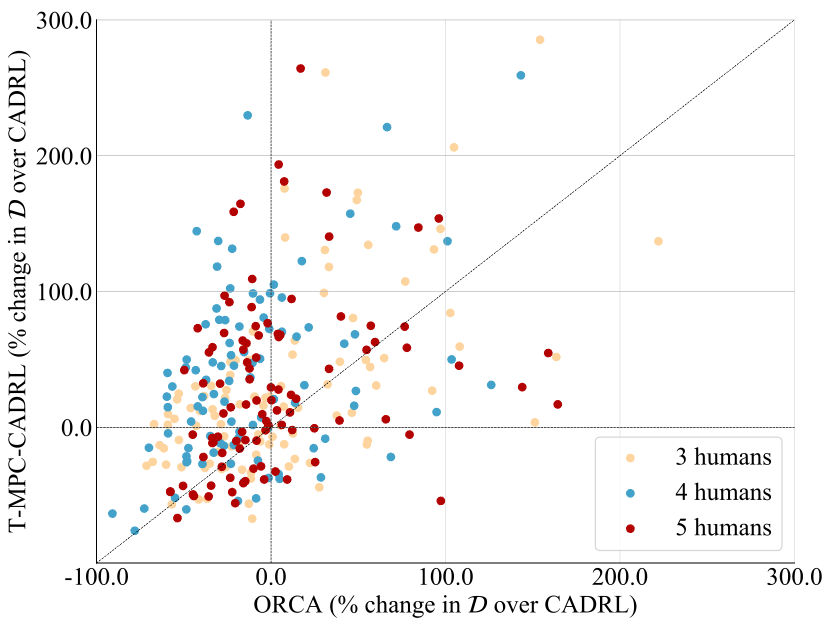}
        \caption{CADRL world.\label{fig:scatter-cadrlworld}}
    \end{subfigure}
    \caption{Scatter plots of $\%$ change in Safety over the world baseline. The plots compare the best-performing T-MPC for each world, i.e., T-MPC-ORCA in~(\subref{fig:scatter-orcaworld}) and T-MPC-CADRL in~(\subref{fig:scatter-cadrlworld}) against CADRL and ORCA respectively. \label{fig:scatterplots-H3}}
\end{figure}

\begin{figure}
    \centering
    \begin{subfigure}{.48\linewidth}
     \centering
         \includegraphics[width = \linewidth]{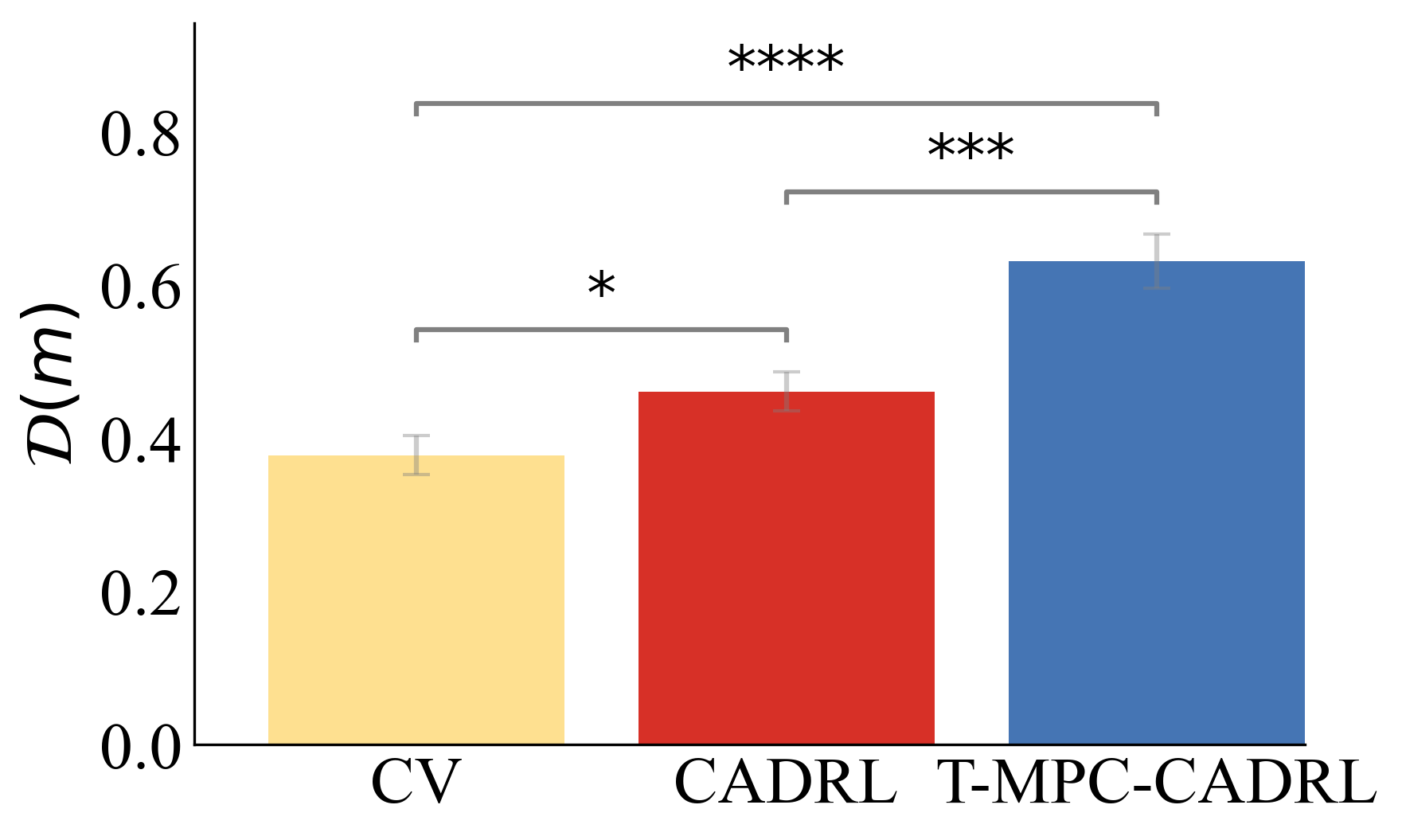}
        \caption{} 
         \label{fig:safety-real}
     \end{subfigure}
    \begin{subfigure}{.48\linewidth}
    \centering
        \includegraphics[width = \linewidth]{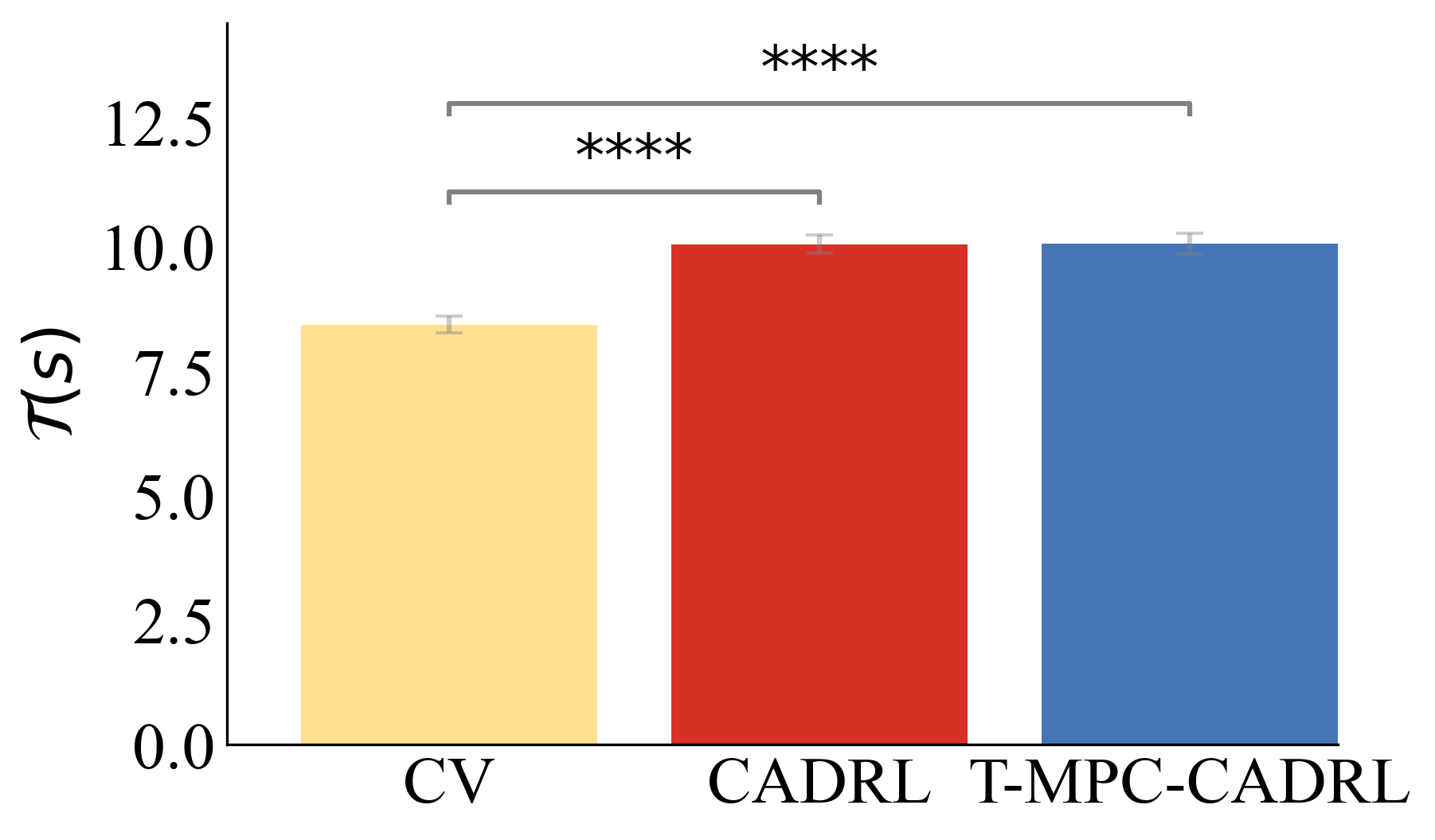}
        \caption{}
        \label{fig:efficiency-real}
    \end{subfigure}
\caption{Robot Safety (\subref{fig:safety-real}) and Time to destination (\subref{fig:efficiency-real}) in real-world experiments. Error bars are $95\%$ confidence intervals. \label{fig:real_world-plots}}
\end{figure}

\tabref{tab:results} contains the controller performance per scenario whereas Figs.~\ref{fig:safety-plots}, \ref{fig:scatterplots-H3}, \ref{fig:real_world-plots} highlight important observations. A video with snippets from our experiments can be found at \href{https://youtu.be/o1o_Rd7MEcg}{this} link.

\textbf{H1}. \figref{fig:safety-plots} depicts the differences between the T-MPC and V-MPC variants in the form of $\%$ improvement over the world baseline (i.e., ORCA or CADRL) in terms of Safety. We see that V-MPC and T-MPC exhibit safer performance when rolling out CADRL or ORCA policies across scenarios in the two simulated worlds. Thus, H1 was supported.

\textbf{H2}. In~\figref{fig:safety-plots}, we see that for identical rollouts, T-MPC is significantly safer than V-MPC across scenarios in the two simulated worlds. Thus, H2 was also supported.

\textbf{H3}. In the simulated worlds, T-MPC variants with CADRL and ORCA rollouts generally perform best in terms of Safety (see~\tabref{tab:results}). \figref{fig:scatterplots-H3} gives a deeper insight containing raw datapoints of comparing T-MPC and CADRL/ORCA: in each world, the best performing T-MPC is the safest. This is even more pronounced in challenging scenarios with 4 or 5 humans. Crucially, as shown in~\figref{fig:real_world-plots}, these trends transfer to the real world; T-MPC-CADRL is safer than CADRL while exhibiting similar efficiency. Thus, we find that H3 was supported.

\section{Discussion}\label{sec:discussion}




T-MPC was safer than baselines (H3), using CV prediction, a very coarse approximation compared to CADRL's collision avoidance mechanism, extracted through training for millions of episodes. It handled robustly different numbers and types of agents ranging from efficient ORCA agents to socially aware CADRL agents in both worlds. Through an ablation study, we showed that T-MPC significantly outperformed alternative competitive V-MPC architectures (H2) which benefit from reactive rollouts (H1). Our investigation underscores the value of integrating a model of passing into robot's decision making: by consistently contributing passing progress, the robot is able to better coordinate safe passages in crowded scenes.

Further, while CADRL and ORCA were more efficient in simulation (see~\tabref{tab:results}), this trend did not transfer to the real world (see~\figref{fig:efficiency-real}). This demonstrates that while T-MPC could be improved for efficiency, its performance is competitive when deployed in realistic settings.





Finally, after the study, users informally confirmed that the differences between policies were noticeable: they found CV to be the least preferred and commended the expressiveness of T-MPC-CADRL but also the predictability of CADRL. While these insights are not statistical, they provide nuance and could inform the design of future large-scale studies~\citep{social-momentum-thri}.

\subsection{Limitations}


\textbf{Evaluation criteria}. Safety and efficiency are necessary but not sufficient attributes for smooth robot performance in crowded domains. Higher-order properties of robot motion such as smoothness or acceleration are also known to influence users' impressions in terms of comfort~\citep{social-momentum-thri}, personality and capabilities~\citep{walker2021corl} and it would be important to account for them.

\textbf{Experimental conditions}. While our evaluation scenarios (\figref{fig:scenarios}) gave rise to challenging human-robot encounters, human motion was cooperative and goal-directed as users were instructed to treat the robot like a human pedestrian. Under such settings, a) our CV-based prediction adapted effectively to the linear segments of human motion; b) our control strategy enabled the robot to expedite pairwise passings, leveraging human cooperation. However, in real-world environments pedestrians are often distracted, rushing, forming groups or changing intentions as they walk, affected by obstacles and the context. Our system would not be able to react appropriately to such motion: CV predictions would not be accurate, and the non-cooperative human response would challenge our winding strategy. Transitioning to such settings would require more expressive models of motion prediction~\citep{rudenko2019-predSurvey} but also the integration of context-aware cost functions. Engineering these settings in the real world would also require new benchmarking experiments that motivate alternative modes of human behavior in an unbiased fashion. In ongoing work, we are developing experimental protocols involving pedestrians with changing intentions, aggressive, and distracted motion. 

\textbf{Hyperparameter tuning}. As detailed in~Sec.\ref{sec:mpc-implementation}, we optimized cost weights for Safety and Time in simulation; we then employed the same weights to the real world without any additional fine tuning as the observed behaviors appeared consistent with simulation. However, we did not perform any optimization over the rollout parametrization; parameters like the number of subgoals, their locations, the rollout discretization and horizon might have influenced performance. In the selection of these parameters, we were guided by the goal of securing a 10Hz control frequency, an empirical standard for real-time response in moderately dynamic environments~\citep{social-momentum-thri,knepper12}. A higher compute budget or a more efficient system implementation could bring performance improvements.

\balance
\bibliography{references}

\begin{thebibliography}{49}
\providecommand{\natexlab}[1]{#1}
\providecommand{\url}[1]{\texttt{#1}}
\expandafter\ifx\csname urlstyle\endcsname\relax
  \providecommand{\doi}[1]{doi: #1}\else
  \providecommand{\doi}{doi: \begingroup \urlstyle{rm}\Url}\fi

\bibitem[Berger(2001)]{Berger2001invariants}
M.~A. Berger.
\newblock Topological invariants in braid theory.
\newblock \emph{Letters in Mathematical Physics}, 55\penalty0 (3):\penalty0
  181--192, 2001.

\bibitem[Bhattacharya et~al.(2011)Bhattacharya, Likhachev, and
  Kumar]{Bhattacharya-RSS-11}
S.~Bhattacharya, M.~Likhachev, and V.~Kumar.
\newblock Identification and representation of homotopy classes of trajectories
  for search-based path planning in 3d.
\newblock In \emph{Proceedings of Robotics: Science and Systems (RSS)}, 2011.

\bibitem[Biswas et~al.(2022)Biswas, Wang, Silvera, Steinfeld, and
  Admoni]{socnavbench}
A.~Biswas, A.~Wang, G.~Silvera, A.~Steinfeld, and H.~Admoni.
\newblock Socnavbench: A grounded simulation testing framework for evaluating
  social navigation.
\newblock \emph{Transactions on Human-Robot Interaction}, 11\penalty0 (3),
  2022.

\bibitem[Brito et~al.(2021)Brito, Everett, How, and Alonso-Mora]{brito-ral21}
B.~Brito, M.~Everett, J.~P. How, and J.~Alonso-Mora.
\newblock Where to go next: Learning a subgoal recommendation policy for
  navigation in dynamic environments.
\newblock \emph{IEEE Robotics and Automation Letters}, 6\penalty0 (3):\penalty0
  4616--4623, 2021.

\bibitem[Cao et~al.(2019)Cao, Trautman, and Iba]{chaocao}
C.~Cao, P.~Trautman, and S.~Iba.
\newblock Dynamic channel: A planning framework for crowd navigation.
\newblock In \emph{Proceedings of the IEEE International Conference on Robotics
  and Automation (ICRA)}, pages 5551--5557, 2019.

\bibitem[Chen et~al.(2019)Chen, Liu, Kreiss, and Alahi]{chen2019crowd}
C.~Chen, Y.~Liu, S.~Kreiss, and A.~Alahi.
\newblock Crowd-robot interaction: Crowd-aware robot navigation with
  attention-based deep reinforcement learning.
\newblock In \emph{Proceedings of the IEEE International Conference on Robotics
  and Automation (ICRA)}, pages 6015--6022, 2019.

\bibitem[{Chen} et~al.(2017){Chen}, {Liu}, {Everett}, and {How}]{chen-icra17}
Y.~F. {Chen}, M.~{Liu}, M.~{Everett}, and J.~P. {How}.
\newblock Decentralized non-communicating multiagent collision avoidance with
  deep reinforcement learning.
\newblock In \emph{Procedings of the IEEE International Conference on Robotics
  and Automation (ICRA)}, pages 285--292, 2017.

\bibitem[Cooper(1990)]{cooper90}
G.~F. Cooper.
\newblock The computational complexity of probabilistic inference using
  bayesian belief networks.
\newblock \emph{Artificial Intelligence}, 42\penalty0 (2):\penalty0 393 -- 405,
  1990.

\bibitem[Everett et~al.(2018)Everett, Chen, and How]{Everett18_IROS}
M.~Everett, Y.~F. Chen, and J.~P. How.
\newblock Motion planning among dynamic, decision-making agents with deep
  reinforcement learning.
\newblock In \emph{Proceedings of the IEEE/RSJ International Conference on
  Intelligent Robots and Systems (IROS)}, pages 3052--3059, 2018.

\bibitem[Feurtey(2000)]{feurtey2000simulating}
F.~Feurtey.
\newblock Simulating the collision avoidance behavior of pedestrians.
\newblock Master's thesis, University of Tokyo, 2000.

\bibitem[Hall(1966)]{hall-1966}
E.~T. Hall.
\newblock \emph{The Hidden Dimension}.
\newblock Anchor Books, 1966.

\bibitem[Honda(2019)]{pathbot}
Honda.
\newblock Honda {P.A.T.H.} {B}ot, 2019.
\newblock URL \url{https://global.honda/innovation/CES/2019/path_bot.html}.

\bibitem[Joosse et~al.(2021)Joosse, Lohse, Berkel, Sardar, and Evers]{joosse21}
M.~Joosse, M.~Lohse, N.~V. Berkel, A.~Sardar, and V.~Evers.
\newblock Making appearances: How robots should approach people.
\newblock \emph{Transactions on Human-Robot Interaction}, 10\penalty0 (1),
  2021.

\bibitem[Kim and Pineau(2016)]{Kim2016}
B.~Kim and J.~Pineau.
\newblock Socially adaptive path planning in human environments using inverse
  reinforcement learning.
\newblock \emph{International Journal of Social Robotics}, 8\penalty0
  (1):\penalty0 51--66, 2016.

\bibitem[Kirby(2010)]{kirby_thesis}
R.~Kirby.
\newblock \emph{Social Robot Navigation}.
\newblock PhD thesis, Carnegie Mellon University, 2010.

\bibitem[Knepper et~al.(2012)Knepper, Srinivasa, and Mason]{knepper12}
R.~A. Knepper, S.~S. Srinivasa, and M.~T. Mason.
\newblock Toward a deeper understanding of motion alternatives via an
  equivalence relation on local paths.
\newblock \emph{The International Journal of Robotics Research}, 31\penalty0
  (2):\penalty0 167--186, 2012.

\bibitem[Koenig and Howard(2004)]{Gazebo}
N.~Koenig and A.~Howard.
\newblock Design and use paradigms for gazebo, an open-source multi-robot
  simulator.
\newblock In \emph{Proceedings of the IEEE/RSJ International Conference on
  Intelligent Robots and Systems (IROS)}, volume~3, pages 2149--2154, 2004.

\bibitem[Kretzschmar et~al.(2016)Kretzschmar, Spies, Sprunk, and
  Burgard]{kretzschmar_ijrr16}
H.~Kretzschmar, M.~Spies, C.~Sprunk, and W.~Burgard.
\newblock Socially compliant mobile robot navigation via inverse reinforcement
  learning.
\newblock \emph{The International Journal of Robotics Research}, 35\penalty0
  (11):\penalty0 1289--1307, 2016.

\bibitem[Lauwers et~al.(2007)Lauwers, Kantor, and Hollis]{oneisenough}
T.~Lauwers, G.~Kantor, and R.~Hollis.
\newblock One is enough!
\newblock In S.~Thrun, R.~Brooks, and H.~Durrant-Whyte, editors, \emph{Robotics
  Research}, pages 327--336. Springer Berlin Heidelberg, 2007.

\bibitem[Lerner et~al.(2007)Lerner, Chrysanthou, and Lischinski]{UCY}
A.~Lerner, Y.~Chrysanthou, and D.~Lischinski.
\newblock Crowds by example.
\newblock \emph{Comput. Graph. Forum}, 26\penalty0 (3):\penalty0 655--664,
  2007.

\bibitem[Liu et~al.(2021)Liu, Chang, Liang, Chakraborty, and
  Driggs-Campbell]{liu2020decentralized}
S.~Liu, P.~Chang, W.~Liang, N.~Chakraborty, and K.~Driggs-Campbell.
\newblock Decentralized structural-rnn for robot crowd navigation with deep
  reinforcement learning.
\newblock In \emph{IEEE International Conference on Robotics and Automation
  (ICRA)}, pages 3517--3524, 2021.

\bibitem[{Lo} et~al.(2019){Lo}, {Yamane}, and {Sugiyama}]{Lo19}
S.~{Lo}, K.~{Yamane}, and K.~{Sugiyama}.
\newblock Perception of pedestrian avoidance strategies of a self-balancing
  mobile robot.
\newblock In \emph{Proceedings of the IEEE/RSJ International Conference on
  Intelligent Robots and Systems (IROS)}, pages 1243--1250, 2019.

\bibitem[Mavrogiannis and Knepper(2021)]{mavrogiannis-hamiltonians}
C.~Mavrogiannis and R.~A. Knepper.
\newblock Hamiltonian coordination primitives for decentralized multiagent
  navigation.
\newblock \emph{The International Journal of Robotics Research}, 40\penalty0
  (10-11):\penalty0 1234--1254, 2021.

\bibitem[{Mavrogiannis} et~al.(2021){Mavrogiannis}, {Baldini}, {Wang}, {Zhao},
  {Trautman}, {Steinfeld}, and {Oh}]{core-challenges2021}
C.~{Mavrogiannis}, F.~{Baldini}, A.~{Wang}, D.~{Zhao}, P.~{Trautman},
  A.~{Steinfeld}, and J.~{Oh}.
\newblock {Core Challenges of Social Robot Navigation: A Survey}.
\newblock \emph{arXiv e-prints}, art. arXiv:2103.05668, Mar. 2021.

\bibitem[Mavrogiannis et~al.(2022)Mavrogiannis, Alves-Oliveira, Thomason, and
  Knepper]{social-momentum-thri}
C.~Mavrogiannis, P.~Alves-Oliveira, W.~Thomason, and R.~A. Knepper.
\newblock Social momentum: Design and evaluation of a framework for socially
  competent robot navigation.
\newblock \emph{Transactions on Human-Robot Interaction}, 11\penalty0 (2),
  2022.

\bibitem[Mavrogiannis and Knepper(2019)]{mavrogiannis-braids}
C.~I. Mavrogiannis and R.~A. Knepper.
\newblock Multi-agent path topology in support of socially competent navigation
  planning.
\newblock \emph{The International Journal of Robotics Research}, 38\penalty0
  (2-3):\penalty0 338–--356, 2019.

\bibitem[Monaci et~al.(2022)Monaci, Aractingi, and Silander]{Monaci-RSS-22}
G.~Monaci, M.~Aractingi, and T.~Silander.
\newblock {DiPCAN: Distilling Privileged Information for Crowd-Aware
  Navigation}.
\newblock In \emph{Proceedings of Robotics: Science and Systems (RSS)}, 2022.

\bibitem[Nishimura et~al.(2020)Nishimura, Ivanovic, Gaidon, Pavone, and
  Schwager]{nishimura2020}
H.~Nishimura, B.~Ivanovic, A.~Gaidon, M.~Pavone, and M.~Schwager.
\newblock Risk-sensitive sequential action control with multi-modal human
  trajectory forecasting for safe crowd-robot interaction.
\newblock In \emph{Proceedings of the IEEE/RSJ International Conference on
  Intelligent Robots and Systems (IROS)}, pages 11205--11212, 2020.

\bibitem[Orthey and Toussaint(2021)]{orthey21}
A.~Orthey and M.~Toussaint.
\newblock Section patterns: Efficiently solving narrow passage problems in
  multilevel motion planning.
\newblock \emph{IEEE Transactions on Robotics}, 37\penalty0 (6):\penalty0
  1891--1905, 2021.

\bibitem[Pellegrini et~al.(2009)Pellegrini, Ess, Schindler, and
  Van~Gool]{PellegriniESG09}
S.~Pellegrini, A.~Ess, K.~Schindler, and L.~Van~Gool.
\newblock You'll never walk alone: Modeling social behavior for multi-target
  tracking.
\newblock In \emph{Proceedings of the International Conference on Computer
  Vision (ICCV)}, pages 261--268, 2009.

\bibitem[Pokorny et~al.(2016)Pokorny, Goldberg, and Kragic]{pokorny16}
F.~T. Pokorny, K.~Goldberg, and D.~Kragic.
\newblock Topological trajectory clustering with relative persistent homology.
\newblock In \emph{Proceedings of the IEEE International Conference on Robotics
  and Automation (ICRA)}, pages 16--23, 2016.

\bibitem[Pérez~D’Arpino et~al.(2021)Pérez~D’Arpino, Liu, Goebel,
  Martín-Martín, and Savarese]{darpino}
C.~Pérez~D’Arpino, C.~Liu, P.~Goebel, R.~Martín-Martín, and S.~Savarese.
\newblock Robot navigation in constrained pedestrian environments using
  reinforcement learning.
\newblock In \emph{Proceedings of the IEEE International Conference on Robotics
  and Automation (ICRA)}, pages 1140--1146, 2021.

\bibitem[Rae et~al.(2013)Rae, Takayama, and Mutlu]{rae13}
I.~Rae, L.~Takayama, and B.~Mutlu.
\newblock The influence of height in robot-mediated communication.
\newblock In \emph{Proceedings of the ACM/IEEE International Conference on
  Human-Robot Interaction (HRI)}, pages 1--8, 2013.

\bibitem[Reinhardt et~al.(2021)Reinhardt, Prasch, and Bengler]{reinhardt21}
J.~Reinhardt, L.~Prasch, and K.~Bengler.
\newblock Back-off: Evaluation of robot motion strategies to facilitate
  human-robot spatial interaction.
\newblock \emph{Transactions on Human-Robot Interaction}, 10\penalty0 (3),
  2021.

\bibitem[Roh et~al.(2020)Roh, Mavrogiannis, Madan, Fox, and
  Srinivasa]{roh2020corl}
J.~Roh, C.~Mavrogiannis, R.~Madan, D.~Fox, and S.~Srinivasa.
\newblock Multimodal trajectory prediction via topological invariance for
  navigation at uncontrolled intersections.
\newblock In \emph{Proceedings of the Conference on Robot Learning}, 2020.

\bibitem[Rudenko et~al.(2020)Rudenko, Palmieri, Herman, Kitani, Gavrila, and
  Arras]{rudenko2019-predSurvey}
A.~Rudenko, L.~Palmieri, M.~Herman, K.~M. Kitani, D.~M. Gavrila, and K.~O.
  Arras.
\newblock Human motion trajectory prediction: a survey.
\newblock \emph{The International Journal of Robotics Research}, 39\penalty0
  (8):\penalty0 895--935, 2020.

\bibitem[Salvini et~al.(2022)Salvini, Paez-Granados, and
  Billard]{salvini-billard}
P.~Salvini, D.~Paez-Granados, and A.~Billard.
\newblock Safety concerns emerging from robots navigating in crowded pedestrian
  areas.
\newblock \emph{International Journal of Social Robotics}, 14:\penalty0
  441–462, 2022.

\bibitem[Salzmann et~al.(2020)Salzmann, Ivanovic, Chakravarty, and
  Pavone]{trajectron}
T.~Salzmann, B.~Ivanovic, P.~Chakravarty, and M.~Pavone.
\newblock Trajectron++: Dynamically-feasible trajectory forecasting with
  heterogeneous data.
\newblock In \emph{Proceedings of the European Conference on Computer Vision
  (ECCV)}, pages 683--700, 2020.

\bibitem[Schöller et~al.(2020)Schöller, Aravantinos, Lay, and
  Knoll]{aravantinos}
C.~Schöller, V.~Aravantinos, F.~Lay, and A.~Knoll.
\newblock What the constant velocity model can teach us about pedestrian motion
  prediction.
\newblock \emph{IEEE Robotics and Automation Letters}, 5\penalty0 (2):\penalty0
  1696--1703, 2020.

\bibitem[Shkurti and Dudek(2017)]{shkurti17}
F.~Shkurti and G.~Dudek.
\newblock Topologically distinct trajectory predictions for probabilistic
  pursuit.
\newblock In \emph{Proceedings of the IEEE/RSJ International Conference on
  Intelligent Robots and Systems (IROS)}, pages 5653--5660, 2017.

\bibitem[Sun et~al.(2021)Sun, Baldini, Trautman, and Murphey]{SunM-RSS-21}
M.~Sun, F.~Baldini, P.~Trautman, and T.~Murphey.
\newblock {Move Beyond Trajectories: Distribution Space Coupling for Crowd
  Navigation}.
\newblock In \emph{Proceedings of Robotics: Science and Systems (RSS)}, 2021.

\bibitem[Trautman et~al.(2015)Trautman, Ma, Murray, and Krause]{trautmanijrr}
P.~Trautman, J.~Ma, R.~M. Murray, and A.~Krause.
\newblock Robot navigation in dense human crowds: Statistical models and
  experimental studies of human-robot cooperation.
\newblock \emph{International Journal of Robotics Research}, 34\penalty0
  (3):\penalty0 335--356, 2015.

\bibitem[Tsoi et~al.(2020)Tsoi, Hussein, Espinoza, Ruiz, and
  V\'{a}zquez]{tsoi20}
N.~Tsoi, M.~Hussein, J.~Espinoza, X.~Ruiz, and M.~V\'{a}zquez.
\newblock Sean: Social environment for autonomous navigation.
\newblock In \emph{Proceedings of the International Conference on Human-Agent
  Interaction}, page 281–283, 2020.

\bibitem[van~den Berg et~al.(2011)van~den Berg, Guy, Lin, and Manocha]{ORCA}
J.~van~den Berg, S.~J. Guy, M.~Lin, and D.~Manocha.
\newblock Reciprocal n-body collision avoidance.
\newblock In \emph{Robotics Research}, pages 3--19. Springer Berlin Heidelberg,
  2011.

\bibitem[Walker et~al.(2021)Walker, Mavrogiannis, Srinivasa, and
  Cakmak]{walker2021corl}
N.~Walker, C.~Mavrogiannis, S.~S. Srinivasa, and M.~Cakmak.
\newblock Influencing behavioral attributions to robot motion during task
  execution.
\newblock In \emph{Proceedings of the Conference on Robot Learning (CoRL)},
  2021.

\bibitem[Wang et~al.(2021)Wang, Mavrogiannis, and Steinfeld]{wang2021corl}
A.~Wang, C.~Mavrogiannis, and A.~Steinfeld.
\newblock Group-based motion prediction for navigation in crowded environments.
\newblock In \emph{Proceedings of the Conference on Robot Learning (CoRL)},
  2021.

\bibitem[Wolfinger(1995)]{Wolfinger95}
N.~H. Wolfinger.
\newblock {Passing Moments: Some Social Dynamics of Pedestrian Interaction}.
\newblock \emph{Journal of Contemporary Ethnography}, 24\penalty0 (3):\penalty0
  323--340, 1995.

\bibitem[Yamane and Kurosu(2020)]{yamane2019}
K.~Yamane and C.~Kurosu.
\newblock Stable balance controller, March 2020.
\newblock US Patent 16/375,111.

\bibitem[Ziebart et~al.(2009)Ziebart, Ratliff, Gallagher, Mertz, Peterson,
  Bagnell, Hebert, Dey, and Srinivasa]{ziebart2009}
B.~Ziebart, N.~Ratliff, G.~Gallagher, C.~Mertz, K.~Peterson, J.~Bagnell,
  M.~Hebert, A.~Dey, and S.~Srinivasa.
\newblock Planning-based prediction for pedestrians.
\newblock In \emph{Proceedings of the IEEE/RSJ International Conference on
  Intelligent Robots and Systems (IROS)}, pages 3931--3936, 2009.

\end{thebibliography}
\bibliographystyle{abbrvnat}


\end{document}